\pdfoutput=1

\documentclass[11pt]{article}

\usepackage[final]{acl}

\usepackage{times}
\usepackage{latexsym}

\usepackage[T1]{fontenc}

\usepackage[utf8]{inputenc}

\usepackage{microtype}

\usepackage{inconsolata}

\usepackage{graphicx}
\usepackage{subcaption}
\usepackage{booktabs}
\usepackage{url}
\usepackage{amsmath}
\usepackage{amssymb}
\usepackage{colortbl}
\usepackage{multirow}
\usepackage{makecell}
\PassOptionsToPackage{prologue,dvipsnames}{xcolor}
\usepackage[dvipsnames]{xcolor}
\definecolor{mylightred}{HTML}{FFE9E8}
\definecolor{mydrakgreen}{HTML}{27967e}
\usepackage{color}
\usepackage{ulem}
\usepackage{paralist}
\usepackage{pifont}
\usepackage{marvosym}

\usepackage{etoc}
\etocdepthtag.toc{mtchapter}
\etocsettagdepth{mtchapter}{subsection}
\etocsettagdepth{mtappendix}{none}

\definecolor{mydarkgreen}{rgb}{0,0.6,0}
\usepackage[most]{tcolorbox}
\usepackage{csquotes}
\usepackage{anyfontsize}
\usepackage{comment}
\usepackage{float}
\usepackage{cuted}
\usepackage{tikz}
\usepackage{listings,multicol}
\usepackage[
    framemethod=tikz,
    skipbelow=\topskip,
    skipabove=\topskip
]{mdframed}
\newtcolorbox[list inside=prompt,auto counter,number within=section]{prompt}[1][]{
    colbacktitle=black!60,
    coltitle=white,
    fontupper=\normalsize,
    boxsep=5pt,
    left=0pt,
    right=0pt,
    top=0pt,
    bottom=0pt,
    boxrule=1pt,
    #1,
}

\title{Eliciting Medical Reasoning with Knowledge-enhanced Data Synthesis: \\ A Semi-Supervised Reinforcement Learning Approach}

\author{
    Haolin~Li$^{1,2}$,
    Shuyang~Jiang$^{1,2}$,
    Ruipeng~Zhang$^{5}$,
    \\
    \bf
    Jiangchao~Yao$^{3,4}$
    Ya~Zhang$^{4,2,6}$,
    Yanfeng~Wang$^{4}$\textsuperscript{\Letter},
    \vspace{1mm}
    \\ 
    \begin{tabular}{c}
    \fontsize{9.5}{9.5}\selectfont
    $^1$College of Computer Science and Artificial Intelligence, Fudan University ~~~
    $^2$Shanghai AI Laboratory \\
    \fontsize{9.5}{9.5}\selectfont
    $^3$CMIC, Shanghai Jiao Tong University ~~~
    $^4$School of Artificial Intelligence, Shanghai Jiao Tong University
    \\
    \fontsize{9.5}{9.5}\selectfont
    $^5$Department of Radiology, Shanghai Sixth People’s Hospital Affiliated to Shanghai Jiao Tong University School of Medicine
    \\
    \fontsize{9.5}{9.5}\selectfont
    $^6$Institute of Artificial Intelligence for Medicine, Shanghai Jiao Tong University School of Medicine
    \vspace{1mm}\\
    \end{tabular}
    \\ 
}

\begin{document}
\maketitle

\renewcommand{\thefootnote}{}
\footnotetext{\Letter: Corresponding author}
\renewcommand{\thefootnote}{\arabic{footnote}} 

\begin{abstract}
While large language models hold promise for complex medical applications, their development is hindered by the scarcity of high-quality reasoning data.
To address this issue, existing approaches typically distill chain-of-thought reasoning traces from large proprietary models via supervised fine-tuning, then conduct reinforcement learning (RL).
These methods exhibit limited improvement on underrepresented domains like rare diseases while incurring substantial costs from generating complex reasoning chains.
To efficiently enhance medical reasoning, we propose MedSSR, a \underline{Med}ical Knowledge-enhanced data \underline{S}ynthesis and \underline{S}emi-supervised \underline{R}einforcement learning framework.
Our framework first employs rare disease knowledge to synthesize distribution‑controllable reasoning questions.
We then utilize the policy model itself to generate high-quality pseudo-labels.
This enables a two‑stage, intrinsic-to-extrinsic training paradigm: self‑supervised RL on the pseudo‑labeled synthetic data, followed by supervised RL on the human-annotated real data.
MedSSR scales model training efficiently without relying on costly trace distillation.
Extensive experiments on Qwen and Llama demonstrate that our method outperforms existing methods across ten medical benchmarks, achieving up to \textbf{+5.93\%} gain on rare-disease tasks.
Our code is available at \url{https://github.com/tdlhl/MedSSR}.

\end{abstract}

\section{Introduction}

\begin{figure}[t]
    \centering
    \includegraphics[width=0.96\linewidth]{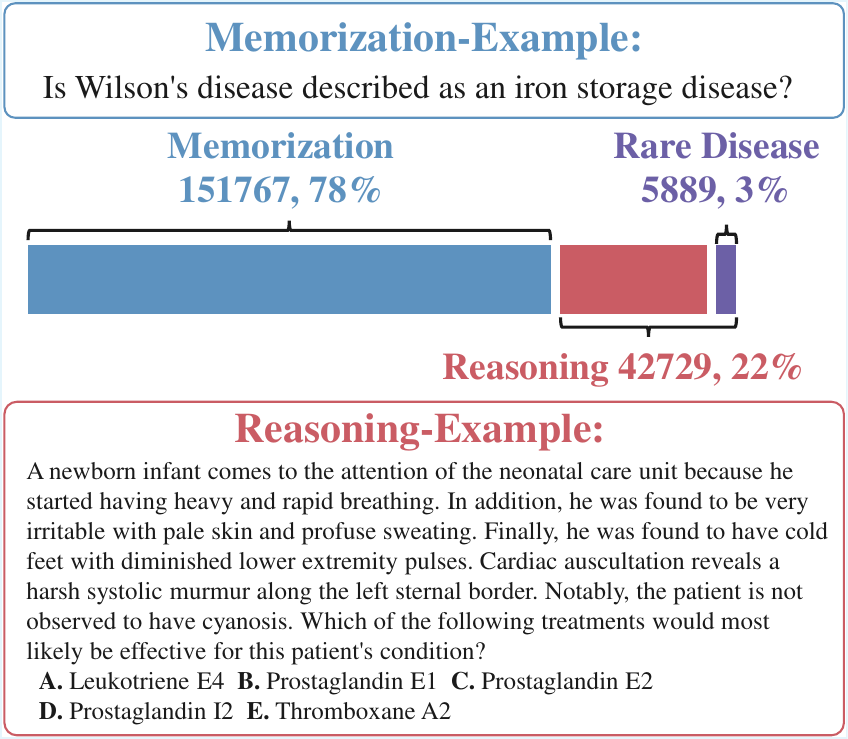}
    \caption{
    Distribution of question types in existing medical datasets. Only 22\% are reasoning-intensive, and just 3\% among them concern rare diseases.
    }
    \label{fig:rare_distribution}
    \vspace{-6pt}
\end{figure}

Large language models (LLMs) have demonstrated remarkable reasoning capabilities~\cite{brown2020language,wei2022chain,achiam2023gpt, chiang2023vicuna}, spurring the development of diverse post-training methods~\cite{ouyang2022training,rafailov2023direct}.
Among them, various methods have been designed to enhance the reasoning ability of LLMs in the medical field, aiming to address complex medical problems~\cite{zhang2023huatuogpt,zhang2024ultramedical}.
However, unlike natural domains where large amounts of reasoning-intensive data are available, downstream domains like healthcare inherently suffer from a scarcity of reasoning-intensive data~\cite{thapa2025disentangling,chen2025heterorag,li2025rad}.
Most existing medical benchmarks consist primarily of memorization-oriented questions, rather than complex reasoning cases.
This issue is especially acute for underrepresented subfields like rare diseases, where labeled data is extremely limited (see Figure~\ref{fig:rare_distribution}).
Due to privacy constraints and the specialized expertise required, acquiring complex medical reasoning data remains exceptionally challenging, yet critical for real-world clinical applications.

\begin{figure*}[t]
    \centering
    \begin{subfigure}{0.6\textwidth}
        \centering
        \includegraphics[width=\linewidth]{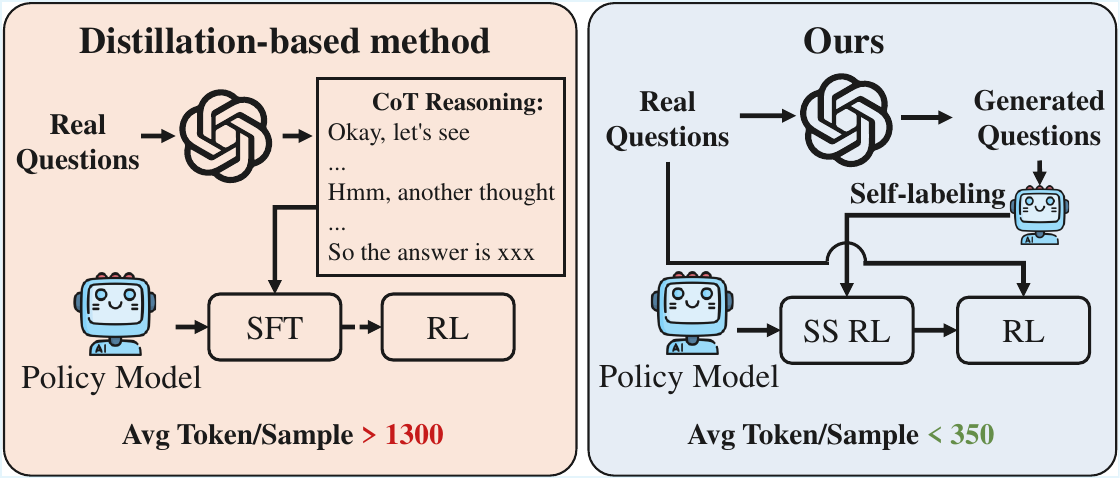}
    \end{subfigure}
    \hfill
    \begin{subfigure}{0.39\textwidth}
        \centering
        \includegraphics[width=\linewidth]{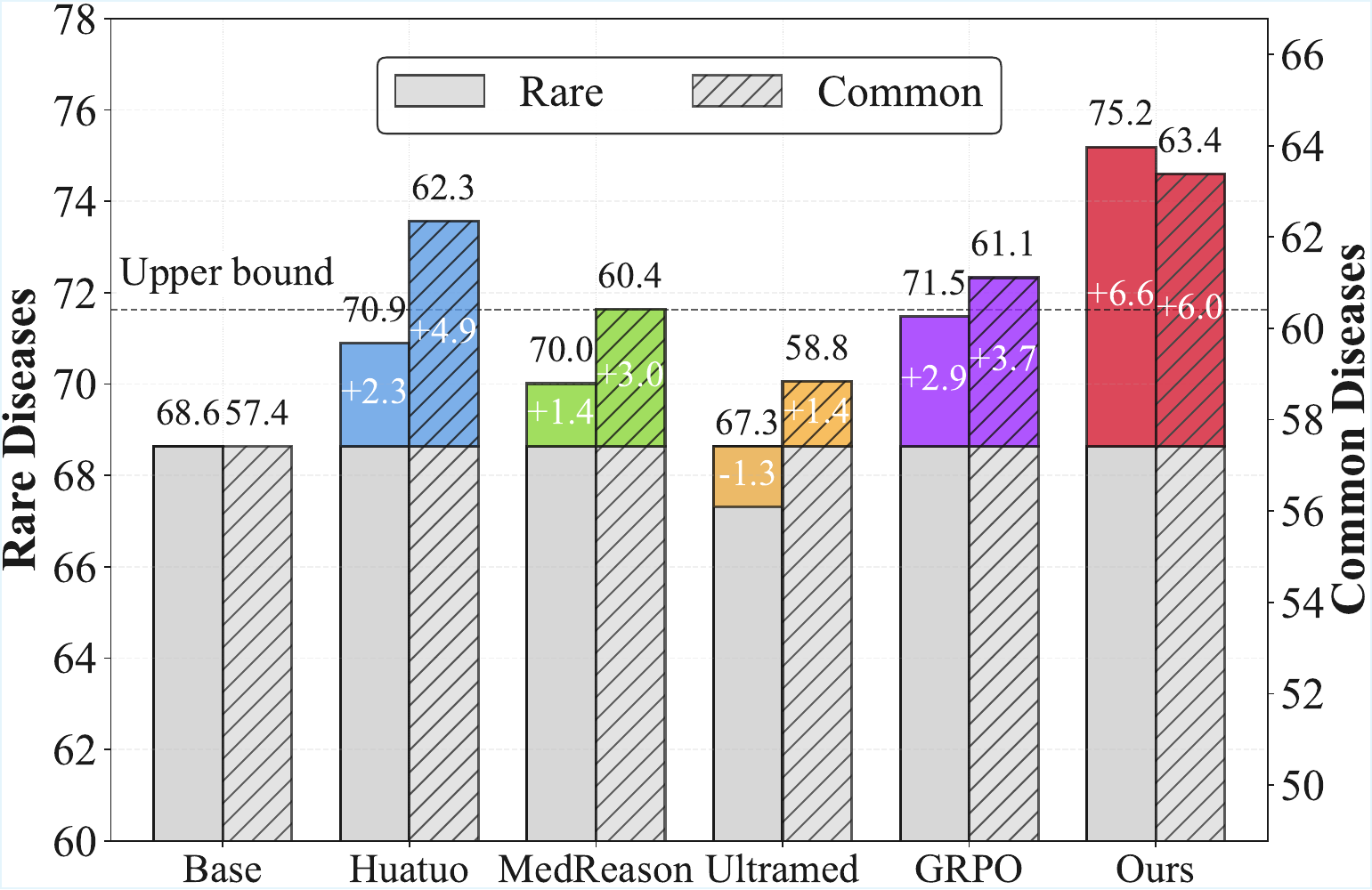}
    \end{subfigure}
    \hfill
    \vspace{-10pt}
    \caption{\textbf{Left}: Comparison of existing methods for medical reasoning.
    ``SS RL'' is short for self-supervised RL.
    ``Avg Token/Sample'' denotes the average number of tokens consumed from an API model to generate one sample.
    \textbf{Right}: Performance improvement comparison based on Llama.
    Prior methods show less improvement in rare diseases than common tasks, while ours significantly breaks the rare disease improvement upper bound of $+3\%$.
    }
    \label{fig:preliminary}
    \vspace{-5pt}
\end{figure*}


Building on this challenge of data scarcity, existing methods for enhancing medical reasoning primarily use large-scale proprietary models to generate chain-of-thought (CoT) reasoning traces~\cite{chen2024huatuogpt}.
The generated traces are then used in supervised fine-tuning to initialize the reasoning capabilities of the policy model, followed by RL training for further improvement.
While effective, these approaches rely on expensive API calls to proprietary models to distill long reasoning chains~\cite{huang2025m1}.
More importantly, they yield limited gains on rare diseases.
As illustrated in Figure~\ref{fig:preliminary}, we find that all existing methods exhibit significantly lower gains on rare diseases compared to common medical tasks.
Even when employing the strong baseline of fully supervised GRPO, performance improvements on rare disease data still fail to surpass a 3\% ceiling.

To efficiently enhance medical reasoning without neglecting data-scarce subdomains, synthetic data generation is a promising alternative.
Recent studies have explored using synthetic data for LLM post-training~\cite{le2022coderl,xu2024wizardlm}.
However, generated data often suffer from low fidelity and factual errors, which are especially unacceptable in medical scenarios~\cite{qin2025scaling,das2025trustworthy}.
Therefore, data synthesis in this domain must incorporate structured medical knowledge to prevent factual errors and ensure clinical rationality.
Furthermore, how to effectively utilize such synthetic data without introducing additional costs or instability remains a challenging problem.

In this paper, we propose MedSSR to efficiently enhance the medical reasoning capabilities of LLMs through the generation and utilization of synthetic data.
Our framework comprises two synergistic components: (1) a medical knowledge-enhanced data synthesis pipeline, and (2) a semi-supervised RL training strategy that jointly leverages synthetic and real data.
MedSSR synthesizes complex medical reasoning questions based on real-world seed data.
Through an optional, on-demand retrieval of rare disease knowledge, it allows explicit control over the proportion of rare disease data.
This enables significant improvement in rare diseases without sacrificing general effectiveness (see Figure~\ref{fig:preliminary}).
To avoid reward hacking, we employ the base model itself to assign pseudo-labels to the synthetic questions via offline majority voting, ensuring the data is well-matched to the model's learning trajectory.
Finally, our training strategy follows an intrinsic-to-extrinsic learning curriculum: we first perform self-supervised RL on pseudo-labeled synthetic data, followed by supervised RL on human-annotated real data.
In summary, our contributions are threefold:
\begin{compactitem}
\item
We identify that existing methods for medical reasoning suffer from suboptimal improvement on rare diseases.
We propose a more effective and efficient paradigm via data synthesis and semi-supervised RL.
\item
We propose MedSSR, a comprehensive framework for leveraging synthetic data, comprising (a) medical knowledge–enhanced data synthesis and (b) a semi-supervised RL training strategy.
This enables controllable data generation and efficient training scale-up.
\item
Extensive experiments on both Qwen3-8B and Llama-3.1-8B-Instruct demonstrate that our approach substantially outperforms baselines and existing medical LLMs, with gains up to \textbf{5.93\%} on rare disease and \textbf{3.91\%} in general.
\end{compactitem}

\section{Related Work}

\subsection{Medical Reasoning Models}
Tremendous efforts have been made to develop medical reasoning models~\cite{huang2025m1}.
HuatuoGPT-O1~\cite{chen2024huatuogpt} distills reasoning traces from GPT-4o via supervised fine-tuning, and then applies a second-stage RL training.
MedReason employs knowledge graphs during CoT generation to improve factual accuracy~\cite{wu2025medreason}.
Another line of work focuses on training powerful process reward models (PRM) to supervise the policy model.
MedS$^3$~\cite{jiang2025meds} and MedPRM~\cite{yun2025med} first generate reasoning paths for PRM training, then use the PRM to supervise the policy model.
While effective, these methods fail to provide considerable gains for rare diseases.
As illustrated in Figure~\ref{fig:preliminary}, MedSSR employs proprietary models only to generate questions instead of long reasoning traces, and trains the policy model via pure RL.
This pipeline bypasses the cost of PRM deployment and trace distillation, achieving higher efficiency while maintaining superior performance on rare diseases.

\subsection{Data Generation via LLMs}
Generating synthetic data has emerged as a powerful strategy to mitigate data scarcity, particularly in domains like healthcare, where collecting real-world data is challenging~\cite{yu2025cot, guo2025synthetic}.
The advent of LLMs has significantly accelerated this paradigm.
Early works like Self-Instruct~\cite{wang2023self} and Alpaca~\cite{taori2023stanford} leveraged LLMs to generate instruction-following data, while later efforts like WizardMath~\cite{luowizardmath} focused on synthesizing complex math problems.
Currently, synthetic data has become an indispensable component for scaling up LLM training~\cite{chen2025self}.
Compared to prior approaches, our method tailors data generation to the medical domain by injecting rare disease knowledge, enhancing the factual accuracy and clinical relevance.

\begin{figure*}[t!]
    \centering
    \includegraphics[width=0.98\linewidth]{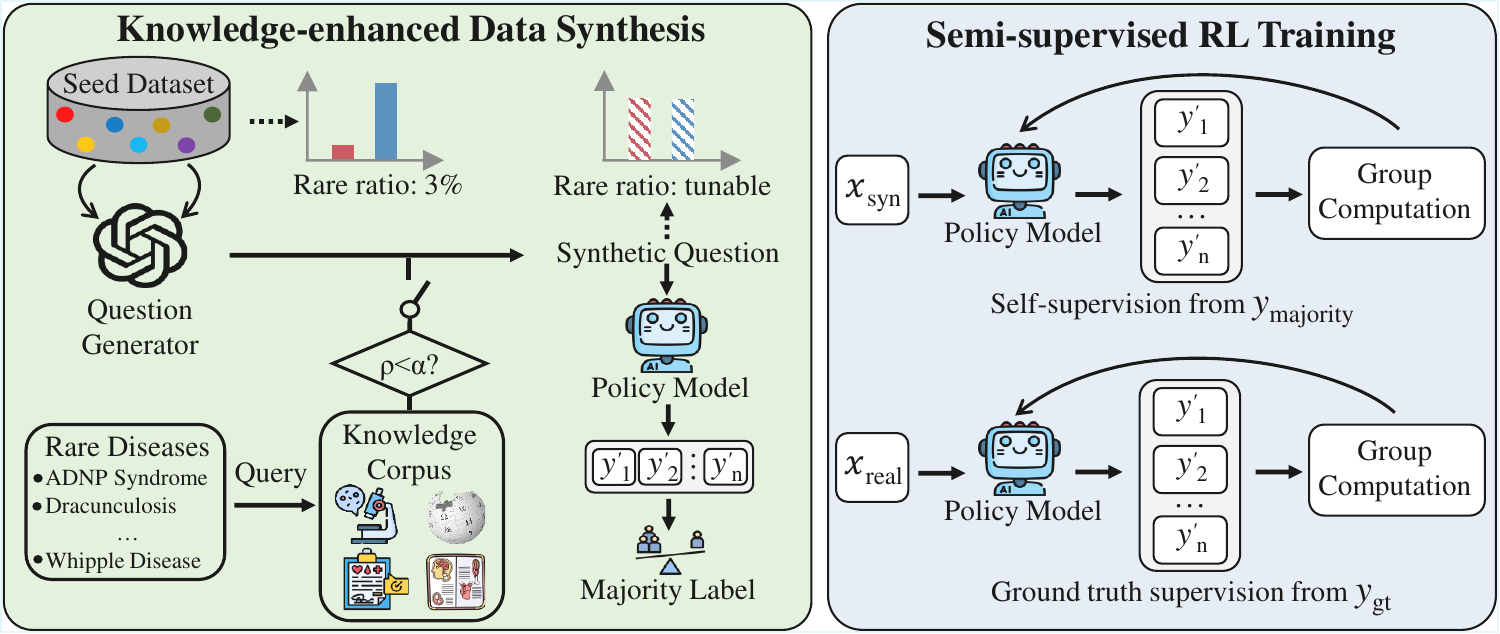}
    \caption{Overview of the proposed method.
    We first synthesize questions with tunable distribution, then train the policy model via a semi-supervised reinforcement learning mechanism.
    Rare disease knowledge injection is triggered when the random value $\rho$ is less than the threshold $\alpha$.
    }
    \label{fig:method}
\end{figure*}

\section{Method}
In this section, we first present the problem formulation in \S~\ref{sec:preliminary}.
Then, we introduce the details of our method in two parts.
\S~\ref{sec:method_synthesis} describes the data synthesis pipeline.
\S~\ref{sec:method_semi} presents the semi-supervised training strategy.
The overall framework of MedSSR is illustrated in Figure~\ref{fig:method}.

\subsection{Preliminary}\label{sec:preliminary}
Let $\mathcal{D} = \{(x, y)\}$ denote the training set, where $x$ is a question and $y$ is the ground-truth answer. 
We aim to optimize an LLM $\pi_\theta$, parameterized by $\theta$, which takes a question $x$ as input and generates a response $y'$ autoregressively.
Typically, the response is a reasoning trace with a final answer $\text{ans}(y')$, where $\text{ans}(\cdot)$ is the function for answer extraction.
The objective is to maximize the verification reward $r(y,y')$, which evaluates the correctness of the answer against the ground truth $y$:
\[
r(y, y') = \mathbb{I}\left[ \text{ans}(y') = y \right],
\]
where $\mathbb{I}[\cdot]$ is the indicator function. The optimization problem can be formulated as:
\begin{multline}\label{eq:base_rl}
\max_{\pi_\theta} \mathbb{E}_{(x, y) \in \mathcal{D},\, y'\sim \pi_\theta(x)} \bigg[ \hat{A}(x, y') \\ 
-\beta \cdot \mathbb{D}_{\text{KL}} \left[ \pi_\theta(y'|x) \,\|\, \pi_{\text{ref}}(y'|x) \right] \bigg],
\end{multline}
where $\hat{A}(x, y')$ denotes the advantage estimated on the reward $r(y,y')$ and the coefficient $\beta$ controls the Kullback-Leibler (KL)~\cite{kullback1951information} divergence between the policy model $\pi_\theta$ and the initial policy $\pi_{ref}$.

\subsection{Knowledge-enhanced Data Synthesis}\label{sec:method_synthesis}
To expand the knowledge coverage of medical LLMs, especially on rare diseases, we introduce a distribution-controllable data synthesis pipeline.


\paragraph{Data Synthesis}\label{sec:enhanced_synthesis}

Given two seed questions $\{x^{s}_1, x^{s}_2\} \subset \mathcal{D}$, we generate new synthetic questions using a powerful proprietary LLM (e.g., GPT-4.1~\cite{hurst2024gpt}), denoted as $\text{LLM}_{\text{gen}}$.
To ground the generation in factual medical knowledge, we augment the question generation pipeline with a comprehensive medical knowledge corpus $\mathcal{P}$, which integrates multiple authoritative sources. 
To precisely control the proportion of rare disease content in the synthetic dataset, we introduce a threshold $\alpha \in [0,1]$.
For each generation step, we sample a random value $\rho \sim \text{Uniform}(0,1)$.
If $\rho < \alpha$, we randomly select a rare disease entity $e$ from a predefined rare disease list $E_{\text{rare}}$ as the query to retrieve the top-$k$ most relevant medical documents from $\mathcal{P}$:
\[
\mathcal{C}(e) = \underset{p \in \mathcal{P}}{\text{Top-}k} \ \text{Sim}(e, p),
\]
where $\text{Sim}(\cdot, \cdot)$ denotes the similarity score, which is computed by the dot product of the embeddings of $e$ and $p$.
These embeddings are encoded by the widely-adopted medical retriever MedCPT~\cite{Jin2023}.
The retrieved rare disease context $\mathcal{C}(e)$ is then concatenated with the seed questions $(x^{s}_1, x^{s}_2)$ to form the input prompt for $\text{LLM}_{\text{gen}}$.

If the random value $\rho \ge \alpha$, no additional disease-specific knowledge is injected.
The synthesis process can then be formulated as:
 \begin{equation}
     x_{\text{syn}} = \begin{cases}
\text{LLM}_{\text{gen}}(x^{s}_1, x^{s}_2, \mathcal{C}(e)), & \text{if } \rho < \alpha, \\
\text{LLM}_{\text{gen}}(x^{s}_1, x^{s}_2), & \text{if } \rho \ge \alpha,
\end{cases}
\end{equation}
where $x_{\text{syn}}$ is the generated question.
Following this procedure, we can obtain a scalable synthetic dataset $\mathcal{D}_{\text{syn}}$ with a tunable focus on rare diseases.

For more details of the rare disease list $E_{\text{rare}}$, knowledge corpus $\mathcal{P}$, and the prompt used for generation, please refer to Appendix~\ref{appendix:method}.

\subsection{Semi-supervised RL training}\label{sec:method_semi}
To utilize the synthetic dataset, we perform self-labeling to avoid labor-intensive human labeling.

\paragraph{Pseudo-label Generation}\label{sec:method_pseudo}

For high-quality data tailored for policy model training, we leverage $\pi_\theta$ itself to generate pseudo-labels.
Given a synthetic question $x_{\text{syn}} \in \mathcal{D}_{\text{syn}}$, we sample $G$ independent responses from the base policy model:
\[
\{y^{1}_{\text{syn}}, y^{2}_{\text{syn}}, \dots, y^{G}_{\text{syn}}\} \sim \pi_\theta(\cdot \mid x_{\text{syn}}).
\]
From each response $y^{i}_{\text{syn}}$, we extract a final answer $a^{i}_{\text{syn}}$. The pseudo-label $y_{\text{majority}}$ is then determined by majority voting over these $G$ answers:
\begin{equation}
y_{\text{majority}} = \arg\max_{a} \sum_{i=1}^{G} \mathbb{I}\left[a^{i}_{\text{syn}} = a\right].
\end{equation}
For samples within the knowledge scope of the base model, which constitute the vast majority of the questions, the majority-voted answers are reliable.
Even when $y_{\text{majority}}$ is incorrect, the reward assignment remains partially valid since other wrong answers still receive zero reward, just as they would under correct supervision.

The resulting labeled dataset $\mathcal{D}_{\text{syn}} = \{(x_{\text{syn}}, y_{\text{majority}})\}$ is then used for self-supervised RL, providing model-aligned intrinsic reward signals.
Crucially, this offline labeling strategy mitigates the reward hacking problem commonly observed in online self-supervised methods, where the policy model collapses to output identical answers across generations to exploit the reward.
In \S~\ref{sec:exp_offline} and Appendix~\ref{appendix:offline_long}, we provide empirical analysis that this offline voting strategy enables long-term stable training for more than 1000 steps.

\paragraph{Self-supervised Training}\label{sec:self_rl}

We employ the Group Relative Policy Optimization (GRPO)~\cite{shao2024deepseekmath} as the algorithm for Eq.~\eqref{eq:base_rl}, which provides a stable and efficient computation for advantages.
The training proceeds in two sequential phases: self-supervised training on synthetic data, followed by supervised training on real data.

\begin{table*}
\centering
\resizebox{\textwidth}{!}{
\begin{tabular}{l|ccccccc|c}
\toprule[1pt]
Model & Symptoms & Causes & Diagnosis & Others & RDs & Treatment & Affects & Avg    \\
\midrule
\multicolumn{9}{l}{\textbf{Large Language Models}} \\
Qwen3-8B       & 68.63    & 72.01  & 57.95     & 66.44  & 70.39             & 52.18     & 68.28   & 65.13  \\
R1-Distill-Llama-8B    & 53.58 & 63.04 & 45.49 & 60.19 & 55.52 & 46.57 & 63.28 & 55.38 \\
\midrule
\multicolumn{9}{l}{\textbf{Medical Large Language Models}} \\
UltraMedical-3.1-8B & 73.53 & 73.01 & 57.28 & 65.28 & 77.47 & 50.83 & 73.79 & 67.31 \\
HuatuoGPT-o1-8B & 74.90 & 78.47 & 61.64 & 70.10 & 79.53 & 52.89 & 78.68 & 70.89 \\
m1-7B-23K& 74.51 & 74.64 & 57.82 & 69.44 & 68.59 & 52.37 & 72.76 & 67.16 \\
MedPRM-8B& 73.14 & 76.00 & 61.79 & 66.20 & 78.45 & 52.37 & 78.97 & 69.56 \\
MedReason-8B & 73.53 & 74.39 & 62.85 & 70.43 & 76.43 & 57.10 & 75.33 & 70.01 \\
\midrule[1pt]
Qwen3-8B-Base& 67.16    & 70.74  & 54.92     & 68.75  & 59.05             & 49.30     & 67.41   & 62.48  \\
~+ Fully-Supervised & 73.24    & 75.41  & 60.65     & 65.97  & 70.39             & 54.41     & 72.24   & 67.47  \\
\rowcolor{gray!12}~+ MedSSR & 79.71& \textbf{80.89}& \textbf{68.19}& \textbf{72.22}& 76.64& 57.51& 78.62& 73.40\\
\rowcolor{gray!12}~\textcolor{mydarkgreen}{$\uparrow$} & \textcolor{mydarkgreen}{6.47} & \textcolor{mydarkgreen}{5.48}   & \textcolor{mydarkgreen}{7.54} & \textcolor{mydarkgreen}{6.25}      & \textcolor{mydarkgreen}{6.25} & \textcolor{mydarkgreen}{3.10}    & \textcolor{mydarkgreen}{6.38}  & \textcolor{mydarkgreen}{5.93}\\
\midrule
Llama-3.1-8B-Instruct& 71.67 & 74.28 & 60.85 & 64.81 & 79.28 & 52.97 & 76.55 & 68.63 \\
~+ Fully-Supervised & 76.27 & 77.72 & 60.18 & 67.13 & 81.09 & 54.55 & 83.45 & 71.48 \\
\rowcolor{gray!12}~+ MedSSR & \textbf{80.00} & 80.30 & 64.96 & 71.99 & \textbf{82.89} & \textbf{59.79} & \textbf{86.38} & \textbf{75.19} \\
\rowcolor{gray!12}~\textcolor{mydarkgreen}{$\uparrow$} & \textcolor{mydarkgreen}{3.73} & \textcolor{mydarkgreen}{2.58} & \textcolor{mydarkgreen}{4.78} & \textcolor{mydarkgreen}{4.86} & \textcolor{mydarkgreen}{1.80} & \textcolor{mydarkgreen}{5.24} & \textcolor{mydarkgreen}{2.93} & \textcolor{mydarkgreen}{3.70} \\
\bottomrule[1pt]
\end{tabular}}
\caption{Performance on rare diseases.
``RDs'' is short for Related Disorders.
The best results are in \textbf{Bold}.
\textcolor{mydarkgreen}{$\uparrow$} denotes the relative improvement of our method compared to the fully-supervised baseline.
Ours is highlighted in gray.
} 
\label{tab:main_rare}
\vspace{-6pt}
\end{table*}

The policy model is first optimized on $\mathcal{D}_{\text{syn}}$ to incentivize its intrinsic reasoning ability.
Given a sample $(x_{\text{syn}}, y_{\text{majority}})$, the reward signal is:
\[
r(y_{\text{majority}}, y_{\text{syn}}) = \mathbb{I}\left[ \text{ans}(y_{\text{syn}}) = y_{\text{majority}} \right].
\]
For each question $x_{\text{syn}}$, we sample a group of $G$ outputs from the current policy model.
The group-wise advantage for each output $y_{\text{syn}}^{i}$ is computed by normalizing its reward against the group statistics:
\begin{equation}
\hat{A}_{\text{syn}}^{i} = \frac{r(y_{\text{majority}}, y_{\text{syn}}^{i}) - \mu_{\text{syn}}}{\sigma_{\text{syn}}},
\end{equation}
where $\mu_{\text{syn}}$ and $\sigma_{\text{syn}}$ are the mean and standard deviation of rewards within the group.
The optimization objective is then formulated as:
\begin{multline}\label{eq:grpo}
\mathcal{J}_{\text{self}}(\theta) = \mathbb{E}_{(x_{\text{syn}}, y_{\text{majority}}) \in \mathcal{D}_{\text{syn}}, \{y_{\text{syn}}^{i}\}_{i=1}^{G} \sim \pi_{\theta_{\text{old}}}(\cdot|x_{\text{syn}})} \\
\Bigg[ \frac{1}{G} \sum_{i=1}^{G} \frac{1}{|y_{\text{syn}}^{i}|} \sum_{t=1}^{|y_{\text{syn}}^{i}|} \Big( \min\bigg[ c_{i,t} \, \hat{A}_{\text{syn}}^{i},\text{clip}(c_{i,t}, \\
1-\epsilon, 1+\epsilon) \, \hat{A}_{\text{syn}}^{i} \bigg] - \beta \, \mathbb{D}_{\text{KL}}\big( \pi_\theta \| \pi_{\text{ref}} \big) \Big) \Bigg],
\end{multline}
where 
\[
c_{i,t} = \frac{\pi_\theta(y_{\text{syn},t}^{i} | x_{\text{syn}}, y_{\text{syn}, <t}^{i})} {\pi_{\theta_{\text{old}}}(y_{\text{syn},t}^{i} | x_{\text{syn}}, y_{\text{syn}, <t}^{i})}
\]
and $\epsilon$ is a clipping hyperparameter.
This phase leverages self-supervised signals to fully exploit the reasoning capacity within the ability boundary of the policy model.

\paragraph{Fully-supervised Training}\label{sec:full_rl}

To further extend the model's capability beyond its intrinsic boundary, we subsequently employ external supervision signals from real-world data.
The training procedure is analogous to the self-supervised phase described in Eq.~\eqref{eq:grpo}, but with the reward computed using ground-truth answers.
Specifically, for a real sample $(x, y) \in \mathcal{D}$, the reward is defined as:
\[
r(y, y^{i}) = \mathbb{I}\left[ \text{ans}(y^{i}) = y \right],
\]
and the group-wise advantage is:
\begin{equation}
\hat{A}^{i} = \frac{r(y, y^{i}) - \mu}{\sigma},
\end{equation}
where $\mu$ and $\sigma$ are the mean and standard deviation of rewards over the $G$ sampled responses for $x$.

The supervised phase further expands the model's reasoning ability based on verified real data.
Together, our intrinsic-to-extrinsic training paradigm maximizes the utility of both synthetic and real data, enabling efficient and effective scaling of medical reasoning models.

\section{Experiments}

\begin{table*}
\resizebox{\textwidth}{!}{
\begin{tabular}{l|ccccccccc|c} 
\toprule[1pt]
Model& BioASQ & MedMCQA & MedQA & MedXpertqa & MMLU  & PubMedQA & NEJM  & Lancet & Medbullets & Avg    \\
\midrule
\multicolumn{11}{l}{\textbf{Large Language Models}} \\
Qwen3-8B       & 75.32  & 62.14   & 66.22 & 15.91      & 83.20 & 48.00    & 61.48 & 63.77  & 48.86      & 58.32  \\
R1-Distill-Llama-8B & 69.82 & 46.42 & 46.60 & 6.28  & 65.24 & 47.30 & 42.08 & 48.36 & 31.66 & 44.86 \\
\midrule
\multicolumn{11}{l}{\textbf{Medical Large Language Models}} \\
UltraMedical-3.1-8B & 81.42 & 60.75 & 64.61 & 13.46 & 75.16 & 58.08 & 64.84 & 60.14 & 51.06 & 58.84 \\
HuatuoGPT-o1-8B & 80.97 & 64.77 & 77.79 & 16.68 & 79.57 & 58.35 & 65.92 & 62.68 & 54.30 & 62.34 \\
m1-7B-23K & 78.55 & 62.23 & 71.84 & 17.10 & 78.94 & 49.90 & 61.28 & 60.26 & 53.82 & 59.32 \\
MedPRM-8B & 79.83 & 58.54 & 63.24 & 15.66 & 76.10 & 58.80 & 63.81 & 60.31 & 48.78 & 58.34 \\
MedReason-8B & 77.34 & 61.56 & 72.00 & 19.14 & 78.38 & \textbf{59.60} & 64.51 & 57.34 & 53.90 & 60.42 \\
\midrule[1pt]
Qwen3-8B-Base  & 76.92 & 58.06 & 62.12 & 13.80  & 79.89 & 49.35 & 60.74 & 60.20 & 46.43 & 56.39  \\
~+ Fully-Supervised & 81.91  & 64.25   & 74.98 & 17.40 & 83.88 & 54.15 & 68.74 & 67.66  & 55.11      & 63.12  \\
\rowcolor{gray!12}~+ MedSSR & \textbf{86.00}& \textbf{67.79}& \textbf{81.14}& \textbf{20.66}& \textbf{86.64}& 54.70& \textbf{71.97} & \textbf{70.94}& \textbf{63.47}& \textbf{67.03}\\
\rowcolor{gray!12}~\textcolor{mydarkgreen}{$\uparrow$} & \textcolor{mydarkgreen}{4.09} & \textcolor{mydarkgreen}{3.54}   & \textcolor{mydarkgreen}{6.16} & \textcolor{mydarkgreen}{3.26}      & \textcolor{mydarkgreen}{2.76} & \textcolor{mydarkgreen}{0.55}    & \textcolor{mydarkgreen}{3.23}  & \textcolor{mydarkgreen}{3.28}   & \textcolor{mydarkgreen}{8.36}      & \textcolor{mydarkgreen}{3.91}   \\
\midrule
Llama-3.1-8B-Instruct& 79.80 & 56.72 & 62.55 & 14.29 & 75.94 & 59.10 & 61.29 & 61.09 & 45.94 & 57.41 \\
~+ Fully-Supervised & 83.41 & 62.50 & 72.25 & 15.46 & 78.67 & 58.65 & 63.85 & 60.86 & 54.38 & 61.11 \\
\rowcolor{gray!12}~+ MedSSR & 84.91 & 64.73 & 74.96 & 17.04 & 80.65 & 58.75 & 66.62 & 65.71 & 57.06 & 63.38 \\
\rowcolor{gray!12}~\textcolor{mydarkgreen}{$\uparrow$} & \textcolor{mydarkgreen}{1.50} & \textcolor{mydarkgreen}{2.23} & \textcolor{mydarkgreen}{2.71} & \textcolor{mydarkgreen}{1.58} & \textcolor{mydarkgreen}{1.98} & \textcolor{mydarkgreen}{0.10} & \textcolor{mydarkgreen}{2.77} & \textcolor{mydarkgreen}{4.85} & \textcolor{mydarkgreen}{2.68} & \textcolor{mydarkgreen}{2.27} \\
\bottomrule[1pt]
\end{tabular}}
\caption{Performance on general medical tasks.
All results are the average of four runs.
} 
\label{tab:main_all}
\vspace{-0.2cm}
\end{table*}

\subsection{Experimental Setup}

\paragraph{Datasets}


We employ 11 widely adopted medical datasets.
Among them, five datasets containing training splits are employed for both training and testing, including: MedMCQA~\cite{Pal2022}, MedQA~\cite{Jin2020}, BioASQ~\cite{Tsatsaronis2015}, HeadQA~\cite{vilares2019head}, and PubMedQA~\cite{Jin2019}.
The remaining benchmarks with only test splits are used exclusively for evaluation, including: MMLU-Med~\cite{Hendrycks2021}, MedXpertqa~\cite{zuomedxpertqa}, Medbullets~\cite{chen2025benchmarking}, NEJM, and Lancet~\cite{thapa2025disentangling}.
For detailed evaluation, we adopt the rare disease focusing benchmark ReDis‑QA~\cite{wang2024assessing} and curate an additional 951 rare‑disease questions from the above benchmarks, forming the RareDis‑Sub set. We categorize its questions into seven types: Symptoms, Causes, Affects, Diagnosis, Related Disorders, Treatment, and Others.
Dataset statistics are detailed in Appendix~\ref{appendix:train_test_data}.

\paragraph{Models and Baselines}
We choose Qwen3-8B-Base~\cite{yang2025qwen3} and Llama-3.1-8B-Instruct~\cite{Dubey2024} as the backbone, and compare our method against the fully-supervised GRPO baseline.
To compare with existing models, we further include baselines in the following categories: (1) General LLMs, including the Qwen3 series, and R1-Distill-Llama-8B~\cite{guo2025deepseek}; (2) Medical LLMs, including HuatuoGPT-o1-8B~\cite{chen2024huatuogpt}, m1-7B-23K~\cite{huang2025m1}, MedReason-8B~\cite{wu2025medreason}, MedPRM-8B~\cite{yun2025med}, and UltraMedical-3.1-8B~\cite{zhang2024ultramedical}.

\paragraph{Training and Evaluation Details}
For RL training, we set the number of rollouts to 8, temperature to 1.0, and response length to 4K.
For the fully-supervised baseline, we carefully selected all the reasoning-heavy samples from existing datasets, resulting in a 43K training set.
For our method, we adopt 43K generated data for self-supervised training and use the same 43K real data for supervised training, resulting in a 1:1 ratio of synthetic data vs. real data.
During evaluation, we set the recommended decoding hyperparameters for each model to ensure optimal performance.
And we report the average accuracy over four runs.
More details are provided in Appendix~\ref{appendix:implementation}.

\subsection{Main Results}\label{sec:exp_main}
By controlling $\alpha$, we adjust the proportion of rare disease content to 25\%, which yields an optimal balance between rare disease and general performance.
As presented in Table~\ref{tab:main_rare} \&~\ref{tab:main_all}, we have the following key findings:

\noindent\textbf{Superiority on Data-scarce Domains:}
MedSSR exhibits significant performance gains on all rare disease tasks.
On Qwen, our method achieves an average improvement of 5.93\% over the supervised baseline, while on Llama, an average gain of 3.70\% is observed, demonstrating the strong generalization across model architectures.
Notably, both models outperform all existing (medical) LLMs.
The Llama-based model exhibits particularly strong initial performance on rare diseases. 
MedSSR can further boost it with 6.56\%, while other Llama-based baselines show marginal improvement.
\begin{table*}[t]
\resizebox{\textwidth}{!}{
\begin{tabular}{c|ccccccccc|c}
\toprule
Stage & BioASQ & MedMCQA & MedQA & MedXpertqa & MMLU & PubMedQA & NEJM & Lancet & Medbullets & Avg \\
\midrule
One-stage & 82.16 & 67.10 & 78.06 & 17.77 & 86.04 & 54.70 & 69.49 & 68.32 & 62.26 & 65.10 \\
\midrule
Reverse-stage-I & 81.91 & 64.25 & 74.98 & 17.40 & 83.88 & 54.15 & 68.74 & 67.66 & 55.11 & 63.12 \\
Reverse-stage-II & 82.83 & 67.48 & 78.67 & 17.27 & 85.56 & 53.30 & 69.36 & 69.66 & 60.55 & 64.96 \\
\midrule
\cellcolor{gray!12}Two-stage-I & \cellcolor{gray!12}82.51 & \cellcolor{gray!12}65.44 & \cellcolor{gray!12}75.61 & \cellcolor{gray!12}17.09 & \cellcolor{gray!12}86.25 & \cellcolor{gray!12}\textbf{55.05} & \cellcolor{gray!12}66.91 & \cellcolor{gray!12}68.87 & \cellcolor{gray!12}55.92 & \cellcolor{gray!12}63.74 \\
\cellcolor{gray!12}Two-stage-II & \cellcolor{gray!12}\textbf{86.00} & \cellcolor{gray!12}\textbf{67.79} & \cellcolor{gray!12}\textbf{81.14} & \cellcolor{gray!12}\textbf{20.66} & \cellcolor{gray!12}\textbf{86.64} & \cellcolor{gray!12}54.70 & \cellcolor{gray!12}\textbf{71.97} & \cellcolor{gray!12}\textbf{70.94} & \cellcolor{gray!12}\textbf{63.47} & \cellcolor{gray!12}\textbf{67.03} \\
\bottomrule
\end{tabular}}
\vspace{-5pt}
\caption{Ablation on training strategy.
``One-stage'' denotes mixed training without separating synthetic/real data.
} 
\label{tab:two_stage}
\vspace{-3pt}
\end{table*}

\noindent\textbf{Effectiveness on General Data:}
As shown in Table 2, our method delivers substantial performance improvements not only on rare diseases but also across a broad spectrum of general medical benchmarks.
For both models, MedSSR consistently surpasses the GRPO baseline across all nine datasets.
Particularly on Qwen, where the supervised baseline already provides a significant gain over 6\%, our method further enhances the average performance by 3.91\%, demonstrating that the benefits of MedSSR extend well beyond data-scarce domains.
By providing richer and more diverse training data, our method can efficiently facilitate training scale-up.
Detailed standard deviations of these results are provided in Appendix~\ref{appendix:main_std}.

\begin{figure}[t]
    \centering
    \includegraphics[width=0.96\linewidth]{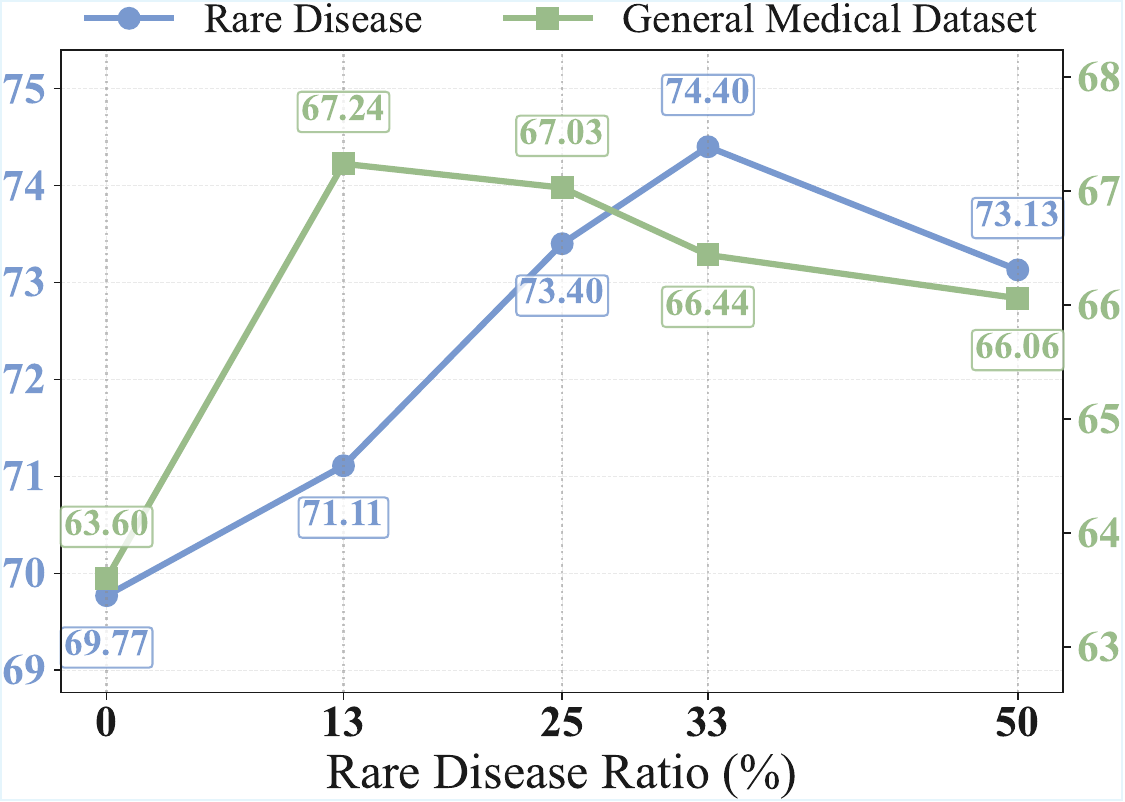}
    \vspace{-5pt}
    \caption{Average performance on rare disease and general datasets across different rare disease ratios.}
    \label{fig:rare_ratio}
    \vspace{-5pt}
\end{figure}

\subsection{Ablation Study}\label{sec:ablation}

\paragraph{Impact of Rare Disease Ratio}\label{sec:rare_ratio}

To investigate the impact of the proportion of rare disease in the training data, we control $\alpha$ to generate synthetic datasets with varying rare disease ratios (0\%, 13\%, 25\%, 33\%, and 50\%).
The 13\% ratio represents the natural distribution obtained when $\alpha$ is always set to 0 (i.e., no rare disease knowledge injection during data synthesis).
The experiments are conducted on Qwen3-8B-Base.

Figure~\ref{fig:rare_ratio} illustrates the average performance on both rare disease and general medical tasks.
As the ratio increases, performance on both sides first increases and then declines.
Notably, general performance reaches its maximum at 13\% and degrades as the ratio increases.
For rare diseases, peak performance occurs at 33\%.
Further increasing the ratio to 50\% declines the abilities on both sides.
The optimal trade-off between rare disease and general performance is achieved at 25\%.
This ratio balances specialization and generalization, maximizing overall utility for real-world applications.

\begin{figure}[t]
    \centering
    \includegraphics[width=\linewidth]{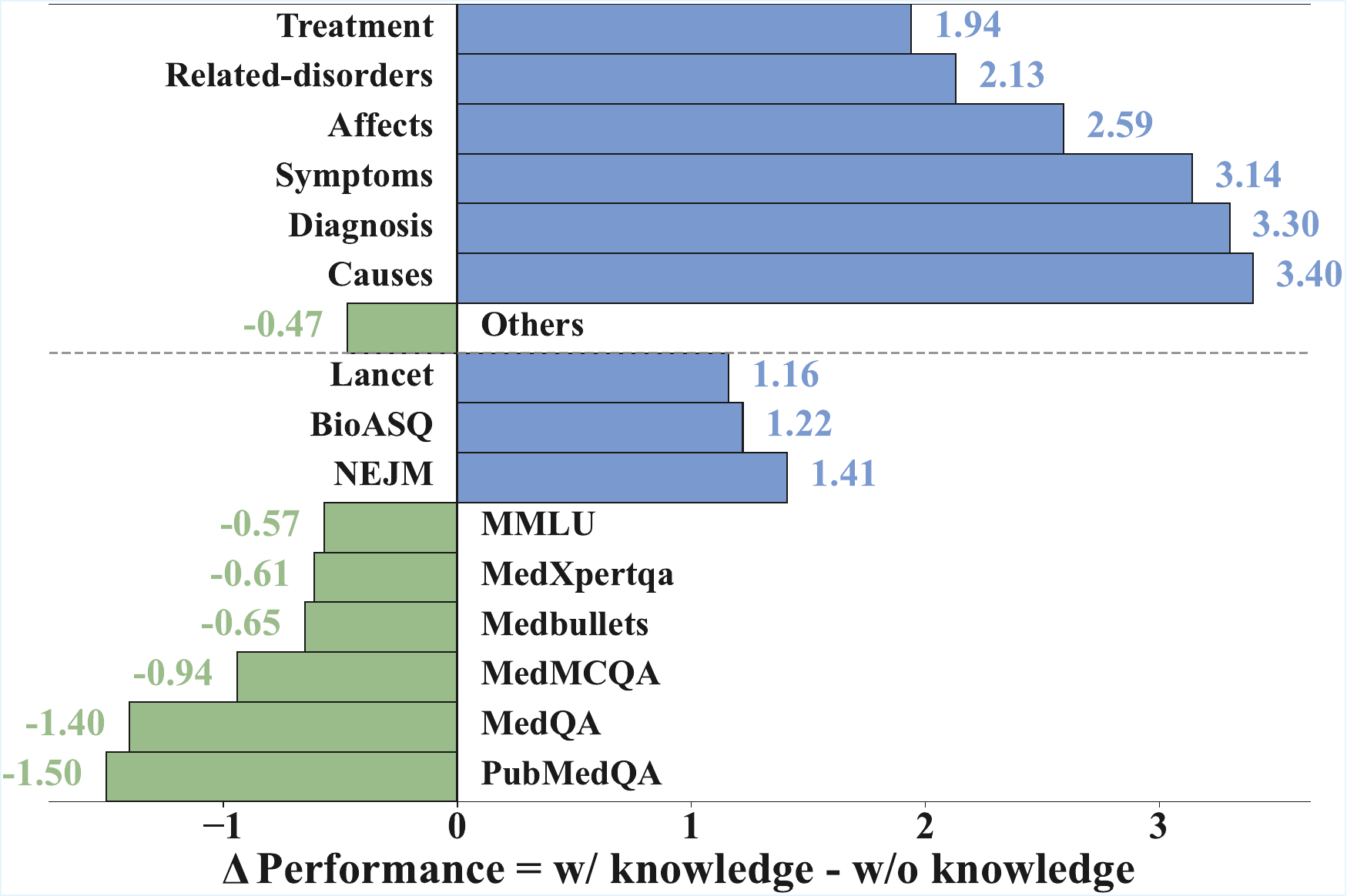}
    \vspace{-14pt}
    \caption{
    Impact of rare disease knowledge injection.
    The upper/lower part denotes rare/general tasks.
    }
    \label{fig:no_inject}
    \vspace{-6pt}
\end{figure}

\paragraph{Impact of Knowledge Injection}\label{sec:wo_injection}
The difference between the 25\% and 13\% ratios directly quantifies the impact of rare disease knowledge injection.
As detailed in Figure~\ref{fig:no_inject}, knowledge injection yields substantial gains across six rare disease categories.
On general benchmarks, performance improves on three datasets while slightly drops on others.
Overall, the large gains on rare diseases (+2.29 in average) outweigh the minimal losses in general (-0.21).
MedSSR enables targeted enhancement in data-scarce domains without compromising overall generalizability.
The ability to precisely control this trade-off makes our method adaptable for various scenarios.
More detailed results of Figure~\ref{fig:rare_ratio} and~\ref{fig:no_inject} on each dataset can be found in Appendix~\ref{appendix:rare_ratio}.

\paragraph{Ablation on the Training Strategy}\label{sec:ablation_semi}

\begin{figure*}[t]
    \centering
    \includegraphics[width=0.98\linewidth]{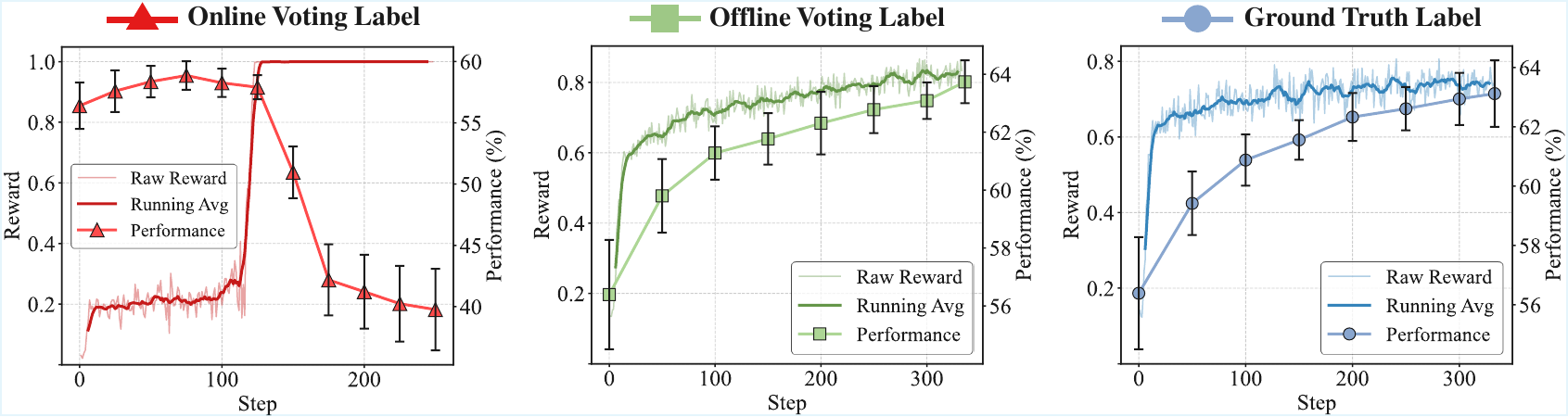}
    \vspace{-7pt}
    \caption{Rewards (left) and performance (right) curves of different labeling strategies across training steps.}
    \label{fig:curve}
    \vspace{-5pt}
\end{figure*}

In this section, we provide an ablation study of our two-stage training strategy.
We compare our method with a single-stage training strategy that does not separate synthetic data and real data.
And a reversed two-stage strategy that first performs supervised training, then self-supervised training.
As presented in Table~\ref{tab:two_stage}, our two‑stage strategy consistently outperforms the single‑stage and the reversed strategy on all benchmarks, validating the necessity of the proposed training paradigm: self‑supervised learning first elicits the model’s intrinsic capability, followed by supervised training to enhance it with external signals.
Notably, the stage-I of our strategy matches or even slightly surpasses supervised training on the same amount of real data, proving the high quality of our generated samples.

\subsection{Further Discussion}\label{sec:exp_further}

\begin{table}[t]
\centering
\resizebox{0.96\linewidth}{!}{
\begin{tabular}{lccc}
\toprule
\textbf{Group} & \textbf{Incorrect} & \textbf{Harmful} & \textbf{Plausibility} \\
\midrule
Senior & 1/200 & 1/200 & 4.86/5 \\
Junior & 0/200 & 0/200 & 4.74/5 \\
\hline
Overall & 0.50\% & 0.50\% & 4.80/5 \\
\bottomrule
\end{tabular}}
\vspace{-5pt}
\caption{Physician evaluation of synthetic samples.}
\label{tab:human_eval}
\vspace{-5pt}
\end{table}

\paragraph{Human Evaluation of Synthetic Data}\label{sec:exp_human} 

To validate the quality of our synthetic data, we invited eight physicians (four senior, four junior) to independently evaluate a subset of 200 samples (labeled by Qwen3-8B-Base).
Each sample was rated for correctness, harmfulness, and plausibility (1‑5 scale). As summarised in Table~\ref{tab:human_eval}, only one sample (0.50\%) was marked incorrect and harmful by senior doctors.
The average plausibility score reached 4.80/5.0.
Both senior and junior groups gave consistently high ratings, confirming the clinical reliability of our generated data.

\paragraph{Offline Labeling Prevents Model Collapse}\label{sec:exp_offline}

In self‑supervised RL, the most straightforward approach is to \textit{online} vote pseudo‑labels at each training step.
However, this strategy is prone to reward hacking.
In this section, we empirically analyze why \textit{offline} labeling is critical for stable training. 

Figure~\ref{fig:curve} presents the training dynamics of: (a) online voting, (b) offline voting, and (c) ground‑truth supervision.
Under online voting, both reward and performance initially rise but soon diverge sharply: the reward rapidly surges to 1.0, while the performance drops substantially, indicating similar outputs across all rollouts.
Similar phenomena also occur in other self-supervised methods based on entropy or other metrics.

In contrast, offline voting yields stable training curves that closely resemble ground‑truth labels: both reward and performance steadily increase.
Decoupling label generation from the training loop can effectively prevent reward hacking.

\begin{figure}[t]
    \centering
    \includegraphics[width=0.95\linewidth]{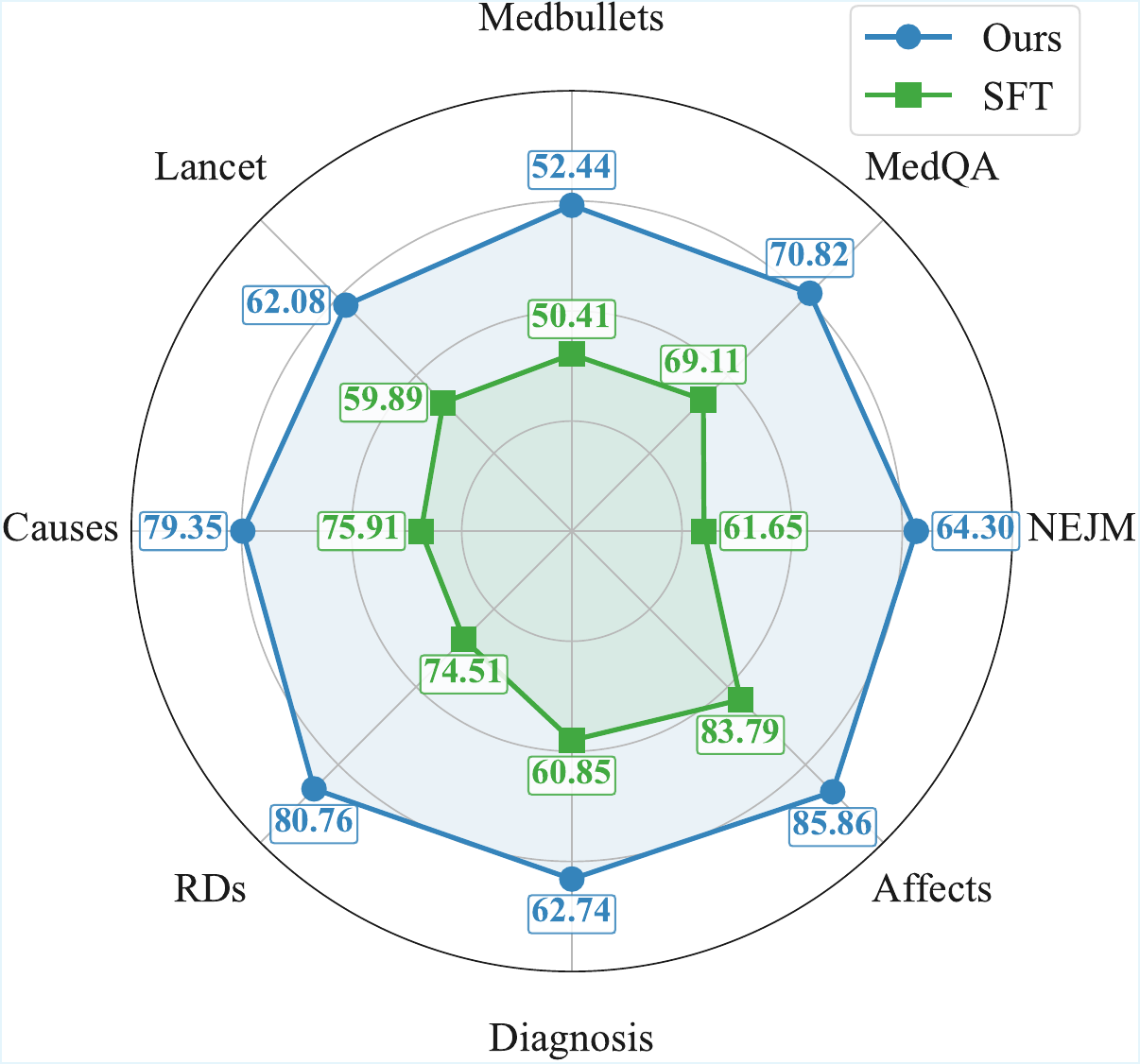}
    \vspace{-7pt}
    \caption{
    Comparison of our question-only synthesis and long reasoning chain synthesis (SFT).
    }
    \label{fig:radar}
    \vspace{-7pt}
\end{figure}

\paragraph{Question-only vs. Reasoning Chain}\label{sec:exp_sft}

To validate that our question-synthesis approach outperforms the prior paradigm of synthesizing entire reasoning chains, we directly compare them on Llama-3.1-8B-Instruct. 
Specifically, ``Ours'' denotes training solely on synthetic questions (43K) via self-supervised RL, and
``SFT'' represents training on the HuatuoGPT-o1 dataset (20K) with 3 epochs via supervised fine-tuning.
As shown in Figure~\ref{fig:radar}, our question‑only approach consistently surpasses the SFT method across all tasks.
These results demonstrate that our method not only reduces generation costs but also delivers superior performance, establishing a more efficient and effective pathway for scaling medical reasoning models.

\section{Conclusion}

In this work, we introduce MedSSR to address the data scarcity and limited improvement in rare diseases when developing medical reasoning models.
Our method includes a medical knowledge-enhanced data synthesis pipeline for generating distribution-controllable questions and a semi-supervised RL training strategy, which first elicits the model's intrinsic ability on synthetic data and then enhances it with ground-truth supervision.
Evaluations across diverse benchmarks and architectures demonstrate the effectiveness of MedSSR.

\section*{Limitations}

Although we have taken concrete steps to mitigate its impact, some limitations remain in our work.
First, due to computational constraints, we did not include experiments on larger LLMs.
Currently, we have tried our best to conduct experiments up to 14B parameter scales.
However, our framework enables efficient scaling.
We believe the core methodology, knowledge-enhanced data synthesis coupled with semi-supervised RL, would generalize effectively to larger models, potentially yielding even greater performance gains as model capacity increases.

Another limitation is that the synthetic data of MedSSR has not been fully validated by human experts.
To mitigate this concern, we have invited eight physicians to assess the factual correctness and clinical plausibility of a subset.
Detailed results and analysis in Appendix~\ref{appendix:human_check} indicate that the majority of generated samples were rated as clinically valid.
This provides preliminary evidence for the reliability of our synthetic data.
Nevertheless, full‑scale expert verification remains desirable for real‑world deployment.



\section*{Ethical Consideration}

The deployment of large language models in healthcare raises ethical considerations regarding data privacy and real-world applicability.
All datasets used in our work are publicly available de-identified medical datasets under licenses that permit research use, and their respective terms are respected in our study.
We ensure that no personally identifiable information is leaked to the question generator (GPT-4.1) or other models that are evaluated.

An advantage of our method is that its enhanced reasoning ability offers more transparent outputs.
By outputting the detailed reasoning process, the model allows medical experts to better understand the response and assess its reliability in clinical contexts.
While our framework demonstrates strong performance on established benchmarks, it is primarily intended for research purposes to advance medical reasoning capabilities.
Any potential real-world deployment would require rigorous clinical validation and expert oversight to ensure safety.

\section*{Acknowledgements}

This work is supported by the National Natural Science Foundation of China (No. 62306178) and STCSM (No. 22DZ2229005), 111 plan (No. BP0719010), and Shanghai Artificial Intelligence Laboratory.

\bibliography{medrag}
\appendix
\clearpage
\appendix 
\hypersetup{pdfborder={0 0 0}}
\etocdepthtag.toc{mtappendix}
\etocsettagdepth{mtchapter}{none}
\etocsettagdepth{mtappendix}{subsection}
\renewcommand{\contentsname}{Appendix}
\tableofcontents 
\hypersetup{pdfborder={0.5 0.5 0.5}}

\section{More Related Work}\label{appendix:related}

\subsection{Reinforcement Learning for LRMs}
Reinforcement learning has become a dominant approach for post-training of large reasoning models (LRMs).
Early approaches employ the PPO algorithm to align models with human preferences using reward models~\cite{schulman2017proximal,ouyang2022training}.
Subsequent RLVR methods simplify the training pipeline by computing rule-based rewards directly from ground-truth answers~\cite{jaech2024openai,lambert2024tulu,guo2025deepseek}. 
GRPO further enhances reasoning capabilities by optimizing group-wise relative policies~\cite{shao2024deepseekmath}.
Another line of research explores self-supervised RL methods, which derive intrinsic rewards from online majority voting answers or response entropy~\cite{shafayat2025can,zhao2025learning}.
However, these approaches often suffer from reward hacking~\cite{prabhudesai2025maximizing,zhang2025right}.
In contrast, MedSSR combines synthetic and real data to form a more stable semi-supervised RL framework, avoiding reward hacking through offline voting on generated data.

\section{Method Details and Verification}\label{appendix:method}

\subsection{Knowledge Corpus}\label{appendix:corpus}

Following prior works~\cite{xiong2024benchmarking,Wu2024a}, we construct the medical knowledge corpus $\mathcal{P}$ from four authoritative sources: (1) PubMed\footnote{\url{https://ftp.ncbi.nlm.nih.gov/pubmed/baseline}} contains the latest biomedical research articles from the PubMed database. (2) Wikipedia includes general medical concepts and descriptions. (3) StatPearls consists of clinical practice summaries and certification exams. (4) Textbooks provide medical textbooks covering diverse domain-specific knowledge.
This multi-source corpus incorporates real-world medical knowledge into the data synthesis process, improving factual accuracy.

\subsection{Rare Disease Sources}\label{appendix:rare_source}

To select samples related to rare diseases in the datasets, we extracted the rare disease list $E_{\text{rare}}$ from four large-scale, certified rare disease databases. The database sources are Orphanet~\cite{weinreich2008orphanet}, Online Mendelian Inheritance in Man~\cite{amberger2015omim}, National Organization for Rare Disorders of the United States, and Compendium of China’s Rare Diseases.
After deduplication, the extracted list $E_{\text{rare}}$ contains a total of 12,445 rare diseases.
$E_{\text{rare}}$ is also employed as the retrieval query in \S~\ref{sec:method_synthesis}.

\subsection{Prompt for Data Synthesis}\label{appendix:syn_prompt}
The input prompt for the question generator $\text{LLM}_{\text{gen}}$ is presented in \S~\ref{appendix:prompt}.
In practice, we adopt GPT-4.1~\cite{hurst2024gpt} to synthesize questions.

\subsection{Expert verification of synthetic data}\label{appendix:human_check}

\begin{table}[htbp]
    \centering
        \resizebox{\linewidth}{!}{
        \begin{tabular}{c|ccccc}
        \toprule[1pt]
            \bf Group & \bf \# Sample & \bf \# Incorrect & \bf \# Harmful & \bf \# Rating & \bf \# SD \\
            \midrule
            Senior & 200 & 1 & 1 & 4.86 & 0.42 \\
            \midrule
            1 & 50 & 1 & 1 & 4.92 & - \\
            2 & 50 & 0 & 0 & 4.78 & - \\
            3 & 48 & 0 & 0 & 4.78 & - \\
            4 & 52 & 0 & 0 & 4.95 & - \\
            \midrule[1pt]
            Junior & 200 & 0 & 0 & 4.74 & 0.54 \\
            \midrule
            1 & 50 & 0 & 0 & 4.98 & - \\
            2 & 51 & 0 & 0 & 4.90 & - \\
            3 & 50 & 0 & 0 & 4.78 & - \\
            4 & 49 & 0 & 0 & 4.30 & - \\
        \bottomrule[1pt]
        \end{tabular}
        }
        \caption{Physician evaluation results of synthetic data.
        ``SD'' denotes the standard deviation of the plausibility rating within the group.
        }
    \label{tab:human_check}
\end{table}

\begin{table}[htbp]
\centering
\resizebox{0.96\linewidth}{!}{
\begin{tabular}{lccc}
\toprule
\textbf{Number} & \textbf{Incorrect} & \textbf{Harmful} & \textbf{Plausibility} \\
\midrule
200 & 0.50\% & 0.50\% & 4.80/5 \\
300 & 1.00\% & 0.67\% & 4.88/5 \\
\hline
500 & 4/500 (0.80\%) & 3/500 (0.60\%) & 4.85/5 \\
\bottomrule
\end{tabular}}
\caption{Expanded Physician evaluation.}
\label{tab:app_human_eval}
\end{table}

The MedSSR synthetic dataset comprises questions generated by the question generator and pseudo‑labels assigned by the policy model itself.
To systematically evaluate the quality of this data against gold standards, we conducted a human evaluation with medical experts in the main content \S~\ref{sec:exp_human}.
Here, we provide a more detailed breakdown of the evaluation results.

We randomly sampled 200 synthetic question–answer pairs (labeled by Qwen3-8B-Base) and invited eight physicians to assess their clinical validity.
All participating physicians were recruited through established clinical‑academic collaborations at our institution.
Each physician received a standard academic honorarium commensurate with their time commitment and expertise level, following our institution's guidelines for professional consultation in research projects.
The doctors were divided into two groups: four junior doctors (3‑5 years of clinical experience; 3 male, 1 female) and four senior doctors (8‑10 years of clinical experience; 2 male, 2 female).
Each group needs to evaluate all samples, with each doctor assessing approximately 50 samples.
They were asked to judge (1) whether the question and its answer were correct, (2) for any incorrect sample, whether it was harmful, and (3) the clinical plausibility of the question on a 1‑5 Likert scale (1 = highly implausible, 5 = highly plausible).

The results are summarised in Table~\ref{tab:human_check}.
Overall, the synthetic data demonstrates high factual accuracy and clinical soundness. Across all 200 questions, only one (0.50\%) was marked as incorrect, and it was flagged as potentially harmful.
The average plausibility score reached 4.80/5.0, indicating that the vast majority of generated questions are considered clinically realistic.
Between the two groups, senior doctors rated the questions slightly higher with lower score variance.
Junior doctors marked zero errors or harmful cases.
These findings suggest that our knowledge‑enhanced synthesis and self-labeling pipeline produces data that is scalable and trustworthy for training medical reasoning models.

Additionally, we expanded the physician evaluation to 500 synthetic samples. 
Three physicians assessed the additional 300 samples under the same criteria described above. 
Results are summarized in Table~\ref{tab:app_human_eval}.
The error rate (0.80\%) and harmful rate (0.60\%) remain extremely low, and the average plausibility score improved to 4.85/5.0. 
These results confirm the robustness of our findings and further validate the high quality of the MedSSR synthetic data.

\section{Experimental Details}\label{appendix:exp_details}

\begin{table}[htbp]
    \centering
        \resizebox{\linewidth}{!}{
        \begin{tabular}{lcccc}
        \toprule[1pt]
            \bf Dataset & \bf \# All & \bf \# Train & \bf \# Test & \bf \# Rare  \\
            \midrule
            BioASQ & 1482& 700 & 782 & 144 \\
            MedQA & 11451 & 10178 & 1273 & 154 \\
            MedMCQA & 184022& 182822 & 1200 & 45 \\
            PubMedQA & 1000& 500 & 500 & 63 \\
            HeadQA & 296 & 296 & - & - \\
            MedXpertqa & 2450 & - & 2450 & 301 \\
            MMLU-Med & 1089& - & 1089 & 57 \\
            NEJM &603 & - & 603 & 65 \\
            Lancet &412 & - & 412 & 70 \\
            Medbullets & 308 & - & 308 & 52 \\
            ReDis-QA & 1360 & - & 1360 & 1360 \\
            \midrule
            \bf Total & 204473  & 194496 & 9977 & 2311 \\
        \bottomrule[1pt]
        \end{tabular}
        }
        \caption{The statistics of the datasets.}
    \label{tab:dataset_stat}
\end{table}


\subsection{Dataset Details}\label{appendix:train_test_data}

The detailed statistics of the datasets adopted in our work are shown in Table~\ref{tab:dataset_stat}.
``\# Rare'' represents the selected test samples related to rare disease that constitute RareDis-Sub.
Detailed information on each dataset is as follows:
\begin{itemize}

\item \textbf{BioASQ}~\cite{Tsatsaronis2015} is a benchmark from an annual biomedical question-answering challenge\footnote{\url{https://www.bioasq.org/}}. We use the official test set of 782 questions from Task B, which require binary yes/no answers.
The remaining 700 samples are used in the training set.
 
\item \textbf{MedQA}~\cite{Jin2020} consists of questions from the United States Medical Licensing Examination (USMLE), covering a wide range of complex clinical scenarios.
The dataset includes both four-option and five-option versions; we use the four-option version and evaluate on the official test set of 1,273 questions.

\item \textbf{MedMCQA}~\cite{Pal2022} is a large-scale dataset sourced from Indian medical entrance exams, spanning over 2,400 healthcare topics.
Due to the absence of labels in the official test set, we select 1,200 questions from the labeled development split as our test set.

\item \textbf{MedXpertqa}~\cite{zuomedxpertqa} is a challenging, expert-level medical benchmark designed for complex clinical reasoning. It includes both text-based QA and multimodal VQA tasks; we use the text-only subset, which contains 2,480 test samples.

\item \textbf{MMLU-Med}~\cite{Hendrycks2021} is the medical subset of the popular MMLU benchmark, covering six biomedical topics: college biology, college medicine, anatomy, clinical knowledge, human genetics, and professional medicine. 
The test set includes 1089 samples.

\item \textbf{PubMedQA}~\cite{Jin2019} is a biomedical question-answering dataset based on PubMed abstracts, originally offering both context-provided and context-free settings. We adopt the context-free version and use the official split.

\item \textbf{HEAD-QA}~\cite{vilares2019head} is a challenging multiple-choice QA benchmark derived from Spanish healthcare specialty exams.
We use the English subset processed by prior works~\cite{chen2024huatuogpt} for training.

\item \textbf{NEJM} and \textbf{Lancet}~\cite{thapa2025disentangling} are real-world medical QA datasets derived from clinical research articles published in medical journals.
Both contain only test samples, and we use the preprocessed versions provided by~\citet{thapa2025disentangling}.

\item \textbf{MedBullets}~\cite{chen2025benchmarking} originates from the same source as MedQA and includes both four-option and five-option multiple-choice questions. We use the five-option version, evaluating on its official test set of 308 questions.

\item \textbf{ReDis-QA}~\cite{wang2024assessing}  is a rare disease–focused benchmark containing 1,360 questions covering 205 distinct rare conditions.
Since it is constructed by extracting samples from existing datasets.
ReDis-QA includes some samples used in the training of our baselines.
Thus, we perform deduplication to ensure no data leakage when testing.

\end{itemize}

To construct the RareDis-Sub benchmark, we first deduplicated the ReDis-QA dataset by removing 189 samples that overlapped with the training sets of other benchmarks, preventing potential data leakage.
Following the rare disease list $E_{\text{rare}}$ derived in \S\ref{appendix:rare_source}, we further screened the remaining nine general medical benchmarks and identified 951 additional samples about rare diseases.
Combining these, we formed the final RareDis-Sub set comprising 2,122 samples.
The samples are categorized into seven categories based on the question focus: Symptoms (255), Causes (552), Affects (145), Related‑disorders (152), Diagnosis (371), Treatment (539), and Others (108).
This categorization enables fine-grained analysis of model performance across diverse aspects of rare disease.

\subsection{Implementation Details}\label{appendix:implementation}

\begin{table}[t]
\centering
\resizebox{0.95\linewidth}{!}{
\begin{tabular}{lcc}
\toprule
Param & Qwen & Llama \\
\midrule
\multicolumn{3}{c}{\textbf{Data Configuration}} \\
\midrule
Global batch size & 128 & 128 \\
PPO mini batch size & 128 & 128 \\
PPO micro batch size & 8 & 8 \\
Max prompt length & 1024 & 1024 \\
Max response length & 4096 & 4096 \\
\midrule
\multicolumn{3}{c}{\textbf{Model Configuration}} \\
\midrule
Algorithm & \texttt{grpo} & \texttt{grpo} \\
Learning rate & 2e-6 & 3e-7 \\
KL loss enabled & True & False\\
KL coefficient & 1e-3 & 0\\
\midrule
\multicolumn{3}{c}{\textbf{Rollout Configuration}} \\
\midrule
Number of rollouts (n) & 8 & 8 \\
Temperature & 1.0 & 1.0 \\
Top-p & 1.0 & 1.0 \\
Top-k & -1 & -1 \\
\bottomrule
\end{tabular}}
\caption{Detailed configuration for RL training.}
\label{tab:rl_config}
\end{table}

\begin{table*}[ht]
\resizebox{\textwidth}{!}{
\begin{tabular}{l|cccccccccc} 
\toprule[1pt]
Model& BioASQ & MedMCQA & MedQA & MedXpertqa & MMLU  & PubMedQA & NEJM  & Lancet & Medbullets & RareDis \\
\midrule
\multicolumn{10}{l}{\textbf{Large Language Models}} \\
Qwen3-8B & 1.90 & 0.23 & 1.33 & 0.36 & 1.08 & 1.19 & 0.68 & 0.54 & 1.31 & 0.41\\
R1-Distill-Llama-8B & 2.08 & 0.88 & 1.31 & 0.64 & 1.61 & 1.63 & 1.59 & 0.88 & 2.51 & 0.40\\
\midrule
\multicolumn{10}{l}{\textbf{Medical Large Language Models}} \\
UltraMedical-3.1-8B & 1.13 & 0.97 & 1.15 & 0.70 & 1.31 & 1.15 & 0.35 & 1.64 & 1.06 & 0.64\\
HuatuoGPT-o1-8B & 1.42 & 0.21 & 0.18 & 0.58 & 1.00 & 0.92 & 1.05 & 1.12 & 0.62 & 0.31\\
m1-7B-23K & 1.96 & 1.52 & 1.51 & 0.76 & 0.93 & 1.30 & 1.37 & 1.82 & 1.53 & 0.32\\
MedPRM-8B & 0.88 & 0.56 & 1.01 & 0.38 & 0.98 & 0.81 & 0.63 & 0.86 & 1.66 & 1.07\\
MedReason-8B & 1.27 & 0.94 & 0.91 & 0.73 & 1.21 & 0.70 & 1.19 & 0.79 & 2.20 & 0.77\\
\midrule[1pt]
Qwen3-8B-Base & 1.83 & 1.08 & 3.79 & 0.74 & 1.62 & 2.01 & 1.13 & 2.53 & 2.28 & 1.52\\
~+ Fully-Supervised & 0.59 & 0.37 & 0.92 & 0.70 & 0.76 & 2.13 & 1.32 & 1.12 & 2.14 & 0.74\\
~+ MedSSR & 0.62 & 0.66 & 0.33 & 0.25 & 0.68 & 0.88 & 0.95 & 1.12 & 1.86 & 0.66\\
\midrule
Llama-3.1-8B-Instruct& 0.43 & 1.00 & 0.28 & 0.71 & 0.78 & 0.77 & 1.24 & 1.96 & 0.84 & 0.55\\
~+ Fully-Supervised & 0.45 & 0.65 & 0.38 & 0.47 & 0.77 & 0.54 & 1.16 & 0.75 & 1.35 & 1.01\\
~+ MedSSR & 0.72 & 0.70 & 0.34 & 0.23 & 0.93 & 0.80 & 0.85 & 0.73 & 1.29 & 0.36 \\
\bottomrule[1pt]
\end{tabular}}
\caption{Detailed standard deviations across all models and benchmarks.
} 
\label{tab:main_all_sd}
\end{table*}

For knowledge retrieval, the top-$k$ is set to $4$.
For the reinforcement learning training, our codes are based on the VeRL framework~\cite{sheng2025hybridflow}.
Since RL training on Llama is prone to collapse, we carefully tuned the hyperparameters for Llama, such as the learning rate and KL loss.
The specific training configurations for the two models are shown in the Table~\ref{tab:rl_config}.
During offline label voting, we set the same configurations as training.

For evaluation, we test all models using the vLLM framework.
We set the recommended decoding hyperparameters for each model to ensure optimal performance.
To ensure answer validity, we apply logit bias in a second decoding pass to avoid invalid outputs.

Additionally, the ``Avg Token/Sample'' values shown in Figure~\ref{fig:preliminary} are computed using the official tokenizers of the corresponding proprietary models.
For our method, we used \texttt{tiktoken}\footnote{\url{https://github.com/openai/tiktoken}} to count tokens in generated questions.
For the SFT baseline, we analyzed the publicly available 23K CoT dataset released by the m1 model\footnote{\url{https://huggingface.co/datasets/UCSC-VLAA/m23k-tokenized}}.
Token counting for DeepSeek-R1 followed its official tokenizer\footnote{\url{https://cdn.deepseek.com/api-docs/deepseek_v3_tokenizer.zip}}.

\section{Additional Experiments}\label{appendix:exp}

\subsection{Standard Deviation of Main Results}\label{appendix:main_std}

\begin{figure}[t]
    \centering
    \includegraphics[width=0.95\linewidth]{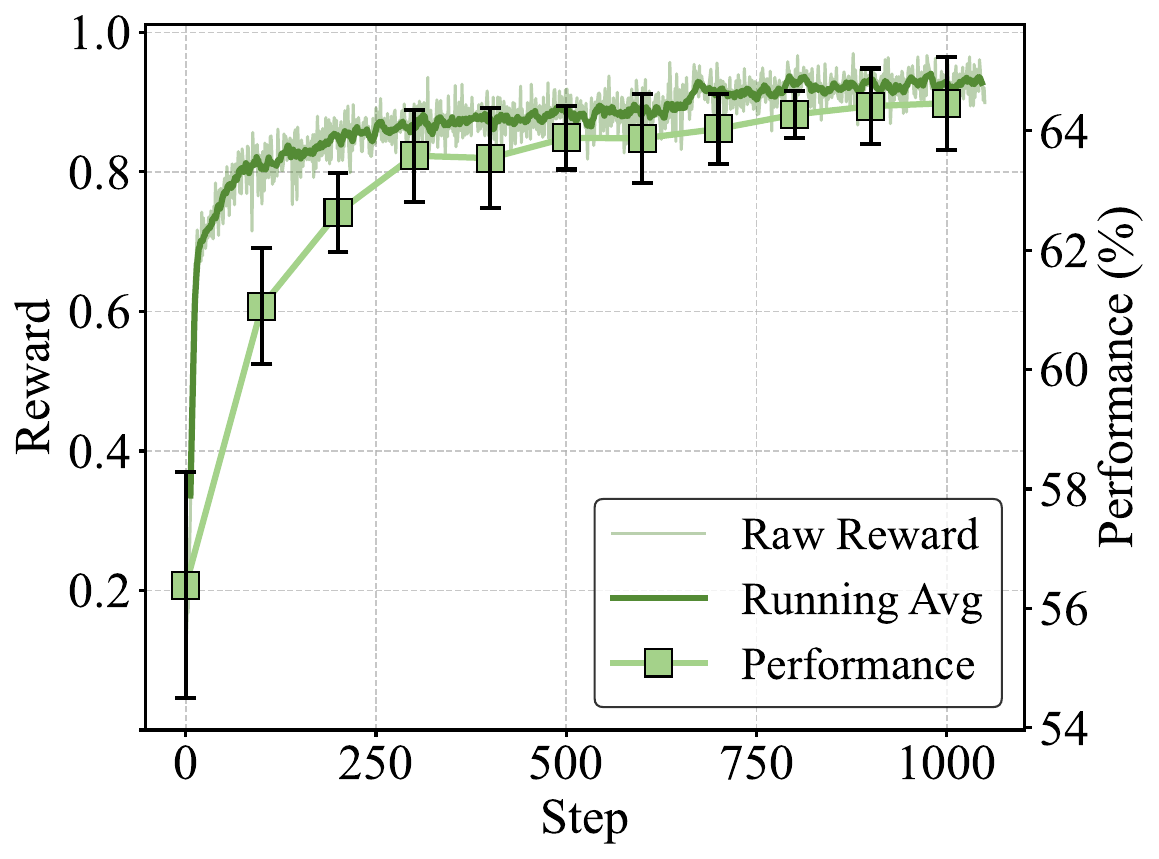}
    \caption{Rewards and performance curves of offline labeling with more than 1K steps.
    We report the average performance on test sets with error bars.
    }
    \label{fig:offline1k}
\end{figure}

\begin{table*}[t]
\centering
\resizebox{0.95\textwidth}{!}{
\begin{tabular}{c|ccccccc|c}
\toprule
Ratio& Symptoms & Causes & Diagnosis & Others & RDs & Treatment & Affects & Avg \\
\midrule
0\%   & 73.73    & 77.76  & 63.07     & 70.37  & 71.22            & 55.84     & 76.38   & 69.77 \\
13\%  & 76.57    & 77.49  & 64.89     & 72.69  & 74.51            & 55.57     & 76.03   & 71.11 \\
25\%  & \textbf{79.71}   & \textbf{80.89}  & 68.19     & 72.22  & 76.64            & 57.51     & 78.62   & 73.40 \\
33\%  & 78.14    & 78.49  & \textbf{68.33}    & \textbf{75.93}  & \textbf{79.77}   & \textbf{59.60}     & \textbf{80.52}   & \textbf{74.40} \\
50\%  & 79.12    & 78.62  & 65.84     & 72.45  & 78.78            & 58.63     & 78.45   & 73.13 \\
\bottomrule
\end{tabular}}
\caption{Performance on RareDis-Sub across different rare disease ratios.
``RDs'' is short for Related Disorders.
} 
\label{tab:rare_ratio_rare}
\end{table*}
\begin{table*}[t]
\resizebox{\textwidth}{!}{
\begin{tabular}{c|ccccccccc|c}
\toprule
Ratio& BioASQ & MedMCQA & MedQA & MedXpertqa & MMLU & PubMedQA & NEJM & Lancet & Medbullets & Avg \\
\midrule
0\%  & 83.47  & 65.89   & 75.69 & 16.40      & 84.53 & 55.20    & 67.50 & 67.59  & 56.09      & 63.60 \\
13\% & 84.78  & \textbf{68.73}   & \textbf{82.54} & \textbf{21.27}      & \textbf{87.21} & \textbf{56.20}    & 70.56 & 69.78  & 64.12      & \textbf{67.24} \\
25\% & \textbf{86.00}  & 67.79   & 81.14 & 20.66      & 86.64 & 54.70    & \textbf{71.97} & \textbf{70.94}  & 63.47      & 67.03 \\
33\% & 83.38  & 68.56   & 81.38 & 20.97      & 86.13 & 53.55    & 71.39 & 67.72  & \textbf{64.86}     & 66.44 \\
50\%  & 84.14  & 67.50   & 80.20 & 19.42      & 86.36 & 56.15    & 69.32 & 66.99  & 64.45      & 66.06 \\
\bottomrule
\end{tabular}}
\caption{Performance on general medical datasets across different rare disease ratios.
} 
\label{tab:rare_ratio_all}
\end{table*}
\begin{table*}[t]
\resizebox{\textwidth}{!}{
\begin{tabular}{cc|ccccccccc|c}
\toprule
Data Scale & Method & BioASQ & MedMCQA & MedQA & MedXpertqa & MMLU & PubMedQA & NEJM & Lancet & Medbullets & Avg \\
\midrule
\multirow{2}{*}{$\times 1$} & Fully-supervised & 81.91 & 64.25 & 74.98 & 17.40 & 83.88 & 54.15 & 68.74 & 67.66 & 55.11 & 63.12 \\
& Self-supervised & 82.51 & 65.44 & 75.61 & 17.09 & 86.25 & \textbf{55.05} & 66.91 & 68.87 & 55.92 & 63.74 \\
\midrule
\multirow{2}{*}{$\times 2$} & Fully-supervised & 81.40 & 66.79 & 77.36 & 16.97 & 84.78 & \textbf{55.05} & 69.03 & 68.14 & 58.60 & 64.24 \\
& MedSSR & \textbf{86.00} & \textbf{67.79} & \textbf{81.14} & \textbf{20.66} & \textbf{86.64} & 54.70 & \textbf{71.97} & \textbf{70.94} & \textbf{63.47} & \textbf{67.03} \\
\bottomrule
\end{tabular}}
\caption{Performance across nine datasets under similar training cost.
We define ``×1 scale'' as the 43K data volume used in prior experiments.
The results of both ``×1 scale'' and ``×2 scale'' experiments are reported for comparison.
} 
\label{tab:86k_all}
\end{table*}

Due to page limitations, we have not presented the standard deviations in our main results in \S~\ref{sec:exp_main}. Here, we provide a detailed breakdown of standard deviations across all models and benchmarks in Table~\ref{tab:main_all_sd}.

Notably, our method exhibits consistently lower standard deviations compared to the fully supervised baselines and the base model on Qwen.
In summary, our method can achieve higher average performance with superior training stability, making it better suited for deployment in safety-critical domains.

\subsection{Offline Voting Enables Long-term Stable Training}\label{appendix:offline_long}
In self-supervised reinforcement learning for LLM post-training, the online strategy of computing internal rewards typical rely on majority voting ~\cite{zuo2025ttrl} or metrics calculated based on the distribution of output~\cite{she2025dupo}.
However, these methods are often prone to reward hacking, leading to model collapse in the early stages of training.
This phenomenon has been discussed in previous works~\cite{jayalath2025compute,liang2025beyond,zhou2025evolution}.
In this work, we adopt a simple offline voting strategy to prevent this.

To further substantiate that offline voting enables sustained and stable training, we extend the self‑supervised training curve beyond 1,000 steps.
As shown in Figure~\ref{fig:offline1k}, both the raw reward and its running average remain steadily increasing and correlated with the actual model performance.
There is no sign of the sharp reward increase or performance collapse in online voting methods (see \S~\ref{sec:exp_offline}).
This long-term stability demonstrates the scalability of our approach for efficiently scaling up RL training, especially for larger models.

\subsection{Impact of Rare Disease Ratio}\label{appendix:rare_ratio}

In this subsection, we provide the full results of the ablation on the rare disease ratio discussed in \S~\ref{sec:rare_ratio}.
The detailed results on RareDis-Sub and general medical benchmarks are presented in Table~\ref{tab:rare_ratio_rare} and Table ~\ref{tab:rare_ratio_all}, respectively.

As discussed in the main text, performance on both rare disease and general tasks follows a consistent trend: it initially improves with increasing rare disease ratio, peaks at an optimal point, and then gradually declines.
The detailed results confirm that a ratio of 25\% strikes the best balance, delivering strong gains on rare disease tasks (e.g., +2.29\% over the 13\% ratio natural distribution) while maintaining competitive performance on general benchmarks. This quantitative breakdown supports our analysis of the trade‑off between specialization and generalization, and justifies the selection of 25\% as the default ratio in our main experiments.

\begin{table*}[t]
\centering
\resizebox{\textwidth}{!}{
\begin{tabular}{l|ccccccc|c}
\toprule
Model& Symptoms & Causes & Diagnosis & Others & RDs & Treatment & Affects & Avg \\
\midrule
Qwen3-1.7B-Base      & 37.35 & 46.65 & 34.03 & 40.05 & 37.17 & 33.77 & 35.86 & 37.84 \\
\rowcolor{gray!12}~+ MedSSR & 49.71 & 55.43 & 44.81 & 48.84 & 49.84 & 40.35 & 47.24 & 48.03 \\
\rowcolor{gray!12}~\textcolor{mydarkgreen}{$\uparrow$} &\textcolor{mydarkgreen}{12.36} & \textcolor{mydarkgreen}{8.78} & \textcolor{mydarkgreen}{10.78} & \textcolor{mydarkgreen}{8.79} & \textcolor{mydarkgreen}{12.67} & \textcolor{mydarkgreen}{6.58} & \textcolor{mydarkgreen}{11.38} & \textcolor{mydarkgreen}{10.19} \\
\midrule
Qwen3-4B-Base& 60.29 & 66.80 & 48.99 & 58.80 & 53.62 & 45.73 & 63.97 & 56.89 \\
\rowcolor{gray!12}~+ MedSSR & 70.69 & 75.14 & 58.69 & 68.29 & 70.23 & 53.76 & 72.41 & 67.03 \\
\rowcolor{gray!12}~\textcolor{mydarkgreen}{$\uparrow$} &\textcolor{mydarkgreen}{10.40} & \textcolor{mydarkgreen}{8.34} & \textcolor{mydarkgreen}{9.70} & \textcolor{mydarkgreen}{9.49} & \textcolor{mydarkgreen}{16.61} & \textcolor{mydarkgreen}{8.03} & \textcolor{mydarkgreen}{8.44} & \textcolor{mydarkgreen}{10.14} \\
\midrule
Qwen3-8B-Base      & 67.16 & 70.74 & 54.92 & 68.75 & 59.05 & 49.30 & 67.41 & 62.48 \\
\rowcolor{gray!12}~+ MedSSR & 79.71 & 80.89 & \textbf{68.19} & 72.22 & 76.64 & 57.51 & 78.62 & 73.40 \\
\rowcolor{gray!12}~\textcolor{mydarkgreen}{$\uparrow$} & \textcolor{mydarkgreen}{12.55} & \textcolor{mydarkgreen}{10.15} & \textcolor{mydarkgreen}{13.27} & \textcolor{mydarkgreen}{3.47} & \textcolor{mydarkgreen}{17.59} & \textcolor{mydarkgreen}{8.21} & \textcolor{mydarkgreen}{11.21} & \textcolor{mydarkgreen}{10.92} \\
\midrule
Qwen3-14B-Base      & 70.67 & 72.05 & 58.69 & 69.83 & 67.89 & 53.87 & 70.03 & 66.15 \\
\rowcolor{gray!12}~+ MedSSR & \textbf{80.39} & \textbf{82.11} & 67.52 & \textbf{75.93} & \textbf{83.55} & \textbf{62.29} & \textbf{81.38} & \textbf{76.17} \\
\rowcolor{gray!12}~\textcolor{mydarkgreen}{$\uparrow$} &\textcolor{mydarkgreen}{9.72} & \textcolor{mydarkgreen}{10.06} & \textcolor{mydarkgreen}{8.83} & \textcolor{mydarkgreen}{6.10} & \textcolor{mydarkgreen}{15.66} & \textcolor{mydarkgreen}{8.42} & \textcolor{mydarkgreen}{11.35} & \textcolor{mydarkgreen}{10.02} \\
\bottomrule
\end{tabular}}
\caption{Performance on RareDis-Sub across different LLM scales.
``RDs'' is short for Related Disorders.
} 
\label{tab:scale_rare}
\end{table*}
\begin{table*}[t]
\resizebox{\textwidth}{!}{
\begin{tabular}{l|ccccccccc|c}
\toprule
Model& BioASQ & MedMCQA & MedQA & MedXpertqa & MMLU & PubMedQA & NEJM & Lancet & Medbullets & Avg \\
\midrule
Qwen3-1.7B-Base      & 56.10 & 41.36 & 39.20 & 11.58 & 53.81 & 45.35 & 37.94 & 32.16 & 30.20 & 38.63 \\
\rowcolor{gray!12}~+ MedSSR & 71.99 & 48.17 & 46.86 & 12.54 & 61.85 & 49.05 & 48.84 & 46.66 & 38.48 & 47.16 \\
\rowcolor{gray!12}~\textcolor{mydarkgreen}{$\uparrow$} & \textcolor{mydarkgreen}{15.89} & \textcolor{mydarkgreen}{6.81} & \textcolor{mydarkgreen}{7.66} & \textcolor{mydarkgreen}{0.96} & \textcolor{mydarkgreen}{8.04} & \textcolor{mydarkgreen}{3.70} & \textcolor{mydarkgreen}{10.90} & \textcolor{mydarkgreen}{14.50} & \textcolor{mydarkgreen}{8.28} & \textcolor{mydarkgreen}{8.53} \\
\midrule
Qwen3-4B-Base      & 76.53 & 52.44 & 55.58 & 11.87 & 74.03 & 50.20 & 54.39 & 52.49 & 41.56 & 52.12 \\
\rowcolor{gray!12}~+ MedSSR & 82.61 & 63.86 & 72.25 & 16.54 & 81.96 & 54.05 & 65.30 & 64.14 & 57.47 & 62.02 \\
\rowcolor{gray!12}~\textcolor{mydarkgreen}{$\uparrow$} &\textcolor{mydarkgreen}{6.08} & \textcolor{mydarkgreen}{11.42} & \textcolor{mydarkgreen}{16.67} & \textcolor{mydarkgreen}{4.67} & \textcolor{mydarkgreen}{7.93} & \textcolor{mydarkgreen}{3.85} & \textcolor{mydarkgreen}{10.91} & \textcolor{mydarkgreen}{11.65} & \textcolor{mydarkgreen}{15.91} & \textcolor{mydarkgreen}{9.90} \\
\midrule
Qwen3-8B-Base      & 76.92 & 58.06 & 62.12 & 13.80 & 79.89 & 49.35 & 60.74 & 60.20 & 46.43 & 56.39 \\
\rowcolor{gray!12}~+ MedSSR & 86.00 & 67.79 & 81.14 & 20.66 & 86.64 & 54.70 & 71.97 & 70.94 & 63.47 & 67.03 \\
\rowcolor{gray!12}~\textcolor{mydarkgreen}{$\uparrow$} &\textcolor{mydarkgreen}{9.08} & \textcolor{mydarkgreen}{9.73} & \textcolor{mydarkgreen}{19.02} & \textcolor{mydarkgreen}{6.86} & \textcolor{mydarkgreen}{6.75} & \textcolor{mydarkgreen}{5.35} & \textcolor{mydarkgreen}{11.23} & \textcolor{mydarkgreen}{10.74} & \textcolor{mydarkgreen}{17.04} & \textcolor{mydarkgreen}{10.64} \\
\midrule
Qwen3-14B-Base      & 75.58 & 62.98 & 72.45 & 18.32 & 82.30 & 47.80 & 63.06 & 60.07 & 55.19 & 59.75 \\
\rowcolor{gray!12}~+ MedSSR & \textbf{86.01} & \textbf{72.92} & \textbf{84.98} & \textbf{24.00} & \textbf{88.98} & \textbf{55.45} & \textbf{75.91} & \textbf{71.60} & \textbf{72.40} & \textbf{70.25} \\
\rowcolor{gray!12}~\textcolor{mydarkgreen}{$\uparrow$} &\textcolor{mydarkgreen}{10.43} & \textcolor{mydarkgreen}{9.94} & \textcolor{mydarkgreen}{12.53} & \textcolor{mydarkgreen}{5.68} & \textcolor{mydarkgreen}{6.68} & \textcolor{mydarkgreen}{7.65} & \textcolor{mydarkgreen}{12.85} & \textcolor{mydarkgreen}{11.53} & \textcolor{mydarkgreen}{17.21} & \textcolor{mydarkgreen}{10.50} \\
\bottomrule
\end{tabular}}
\caption{Performance on general medical datasets across different LLM scales.
} 
\label{tab:scale_all}
\end{table*}

\subsection{Comparison under Similar Cost}\label{appendix:fair_comparison}

In the main experiments, we employed a 1:1 ratio of synthetic to real data, resulting in our semi‑supervised method using twice the data volume of the fully‑supervised baseline.
To ensure a fair comparison under equivalent training cost, we extend the fully‑supervised baseline to the same data scale.
Specifically, we add an additional 43K real samples from the rest of the collected training data (195K in Table~\ref{tab:dataset_stat}), resulting in an 86K fully‑supervised dataset.

The results are presented in Table~\ref{tab:86k_all}.
Even when doubling the real training data, the fully‑supervised baseline achieves only a marginal average improvement (+1.12\% over the 43K baseline).
The additional 43K real samples contribute little to reasoning performance, likely because they are less suitable for complex reasoning tasks than the curated 43K reasoning samples used in previous experiments.
In contrast, MedSSR with the same total data budget (43K synthetic + 43K real) delivers a substantially larger gain of +3.91\% over the 43K fully‑supervised baseline.
This gap underscores that our synthetic data is not merely “more data” but higher‑quality, reasoning‑targeted data, effectively compensating for the scarcity of reasoning samples in existing medical corpora.
In general, our method enables efficient scaling without requiring additional human annotation or costly data collection.

\subsection{Generalizability across Different Scales}\label{sec:ablation_scale}

To evaluate the scalability and generalizability of our method, we conduct experiments on three scales of the Qwen3 model series (1.7B, 4B, 8B, and 14B).
Results on the rare diseases and on the general benchmarks are summarized in Table~\ref{tab:scale_rare} and Table~\ref{tab:scale_all}, respectively.

Our framework consistently delivers substantial gains across all model sizes.
On rare‑disease tasks, the average improvement exceeds 10\% at every scale.
This demonstrates that our method is particularly effective at amplifying performance in data‑scarce domains, regardless of the base model’s initial capacity.
On general medical benchmarks, the absolute improvement grows noticeably as the model scale increases: from +8.5\% at 1.7B to +10.6\% at 8B. This trend indicates that larger models may benefit more from our synthetic data and semi‑supervised training, likely because they possess a stronger latent knowledge base that can be elicited through post-training.
Overall, these results confirm that MedSSR is not tailored to a specific model size; rather, it provides a scalable and model‑agnostic framework for enhancing medical reasoning, with particularly strong performance in data-scarse subdomains.


\section{Case Studies}\label{appendix:study_case}

\subsection{Synthetic Examples}\label{appendix:case}

\begin{table*}[htbp]
\centering
\resizebox{\linewidth}{!}{
    \begin{tabular}{p{2.5cm}p{20.5cm}}
       \toprule[1pt]
        \multicolumn{2}{c}{\bf Case Study w/ rare disease} \\
       \midrule
       \bf Seed \newline Question 1 & An army recruit, smoker and 6 months into training staed complaining of pain at postero medial aspect of both legs. There was acute point tenderness and the pain was aggravated on physical activity. The most likely diagnosis is \newline A. Bearger's disease B. Gout C. Lumbar canal stenosis D. Stress fracture\\
       \midrule
       \bf Seed \newline Question 2 & A patient presenting with haemoptysis and renal failure with antibasement membrane antibodies has -\newline A. Good pasture's B. Wegener's C. Churg Strauss D. Henoch-schonlein purpura\\
       \midrule
       \bf Rare Disease & \textcolor{purple}{Worster-Drought syndrome/Suprabulbar paresis, congenital}\\
       \midrule
       \bf Retrieved Knowledge & \textcolor{purple}{Worster-Drought syndrome} is a form of congenital suprabulbar paresis that occurs in some children with cerebral palsy. It is caused by inadequate development of the corticobulbar tracts and causes problems with the mouth and tongue including impaired swallowing. A similar syndrome in adults is called anterior opercular syndrome. A 1986 study of a family in which multiple members had \textcolor{purple}{Worster-Drought syndrome} suggested it might be hereditary. A 2000 review of cases classified \textcolor{purple}{Worster-Drought Syndrome} as a form of cerebral palsy, caused by early damage to the brain, but identified no obvious causes during gestation or birth and found some families with a history of the condition. The syndrome was named after Cecil Charles \textcolor{purple}{Worster-Drought}, the doctor who described it in 1956.\\
       \midrule
       \bf Synthetic \newline Question &
       A 5-year-old child presents with significant speech delay, difficulty swallowing since infancy, and drooling. Neurological examination reveals impaired movement of the tongue and soft palate, with no limb weakness. The child’s mother reports similar symptoms in her own childhood. MRI shows underdevelopment of the corticobulbar tracts. Which of the following is the most likely diagnosis? \newline A. Moebius syndrome  B. Spinal muscular atrophy (bulbar form) \newline C. \textcolor{purple}{Worster-Drought syndrome}  D. Anterior opercular syndrome (acquired suprabulbar palsy)\\
       \midrule 
       \bf Majority \newline Answer \newline Num: 5/8 & 
       To determine the most likely diagnosis for this 5-year-old child, let's go through the information step-by-step.\newline \#\#\# Step-by-Step Analysis:\newline1. **Clinical Presentation**:  - Significant speech delay  - Difficulty swallowing since infancy  - Drooling  - Impaired movement of the tongue and soft palate  - No limb weakness. 2. **Family History**:  - Similar symptoms in the mother\u2019s childhood. 3. **Neurological Examination**:  - Impaired movement of the tongue and soft palate  - No limb weakness. 4. **Imaging Findings**:  - MRI shows underdevelopment of the corticobulbar tracts. \newline\#\#\# Differential Diagnosis: 1. **Moebius Syndrome (A)**:  - Characterized by congenital facial paralysis (including inability to move the eyes laterally and the mouth) and sometimes other cranial nerve abnormalities.  - Often associated with feeding difficulties, facial asymmetry, and small eyes.  - Does not typically involve limb weakness or underdevelopment of the corticobulbar tracts. 2. **Spinal Muscular Atrophy (Bulbar Form) (B)**:  - Typically presents with muscle weakness and wasting, particularly in the bulbar muscles (speech, swallowing, and respiratory muscles).  - Often has a progressive course.  - Usually involves limb weakness due to lower motor neuron involvement. 3. **\textcolor{purple}{Worster-Drought Syndrome} (C)**:  - Also known as congenital central hypoventilation syndrome.  - Characterized by underdevelopment of the brainstem, particularly the pontine respiratory centers.  - May present with respiratory issues, but not typically with speech delay or swallowing difficulties unless it affects the cranial nerves.  - Usually involves other neurological symptoms like poor sucking and feeding. 4. **Anterior Opercular Syndrome (D)**:  - Also known as acquired suprabulbar palsy.  - Caused by lesions affecting the corticobulbar tracts, leading to weakness or paralysis of the muscles supplied by the lower motor neurons (e.g., facial, tongue, and pharyngeal muscles).  - Typically presents in adults, not children.  - Associated with spasticity and hyperreflexia due to upper motor neuron involvement.\newline \#\#\# Final Considerations: - The combination of significant speech delay, difficulty swallowing since infancy, drooling, impaired movement of the tongue and soft palate, and underdevelopment of the corticobulbar tracts suggests a congenital condition affecting the brainstem and cranial nerves. - The presence of similar symptoms in the mother\u2019s childhood suggests a familial pattern, which is characteristic of \textcolor{purple}{Worster-Drought syndrome} or Moebius syndrome. Given the specific combination of symptoms and imaging findings, **\textcolor{purple}{Worster-Drought syndrome}** (also known as congenital central hypoventilation syndrome with associated craniofacial anomalies) is the most likely diagnosis. This condition is characterized by underdevelopment of the brainstem, which would explain the underdevelopment of the corticobulbar tracts and the associated cranial nerve abnormalities.\newline \#\#\# Final Answer: The final answer is C.\\
       \midrule
       \bf Minority \newline Answer \newline Num: 3/8 & 
       \# Step-by-step reasoning: To determine the most likely diagnosis for the child, let's break down the information step-by-step.\newline\#\#\# Step 1: Clinical Presentation 1. **Speech delay**: Significant speech delay. 2. **Difficulty swallowing**: Since infancy. 3. **Drooling**: Present. 4. **Neurological examination**: Impaired movement of the tongue and soft palate, no limb weakness. 5. **Family history**: Mother reports similar symptoms in her childhood.\newline\#\#\# Step 2: MRI Findings - **Underdevelopment of the corticobulbar tracts**: This suggests involvement of the motor pathways controlling the muscles of the face, tongue, and pharynx.\newline\#\#\# Step 3: Differential Diagnosis 1. **Moebius syndrome (A)**:  - Characterized by facial paralysis (including inability to move the eyes laterally) and impairment of tongue movement.  - Often associated with feeding difficulties and speech problems.  - Typically congenital, and there may be other anomalies like clubfoot or heart defects.  - No limb weakness. 2. **Spinal muscular atrophy (bulbar form) (B)**:  - Characterized by muscle weakness and wasting, primarily affecting the bulbar muscles (those involved in speech, swallowing, and breathing).  - Often presents with respiratory distress in infancy.  - Progressive muscle weakness, including limbs, is a key feature.  - No specific mention of congenital facial paralysis or family history in this case. 3. **\textcolor{purple}{Worster-Drought syndrome} (C)**:  - Also known as congenital central hypoventilation syndrome (CCHS).  - Characterized by congenital abnormalities of the autonomic nervous system, leading to hypoventilation during sleep.  - Often associated with Hirschsprung disease and Hirschsprung-associated intestinal ganglioneuropathy.  - May have facial paralysis and tongue movement issues, but not typically speech delay or drooling alone.  - MRI findings are not specific for this condition. 4. **Anterior opercular syndrome (acquired suprabulbar palsy) (D)**:  - Characterized by contralateral upper motor neuron (UMN) signs of the face (e.g., central facial palsy), contralateral limb weakness, and contralateral sensory loss.  - Typically due to stroke or other acquired causes.  - Not congenital, and there is usually limb involvement.  - No family history of similar symptoms.\newline\#\#\# Step 4: Family History - The mother reporting similar symptoms suggests a genetic or congenital condition rather than an acquired one.\newline\#\#\# Step 5: MRI Findings - Underdevelopment of the corticobulbar tracts is consistent with congenital conditions affecting these pathways, such as Moebius syndrome or other congenital disorders.\newline\#\#\# Step 6: Final Diagnosis Given the clinical presentation of speech delay, difficulty swallowing since infancy, drooling, impaired tongue and soft palate movement, no limb weakness, and a family history of similar symptoms, the most likely diagnosis is Moebius syndrome.\newline\#\#\# Final Answer The final answer is A.\\
       \bottomrule[1pt]
    \end{tabular}
}
\caption{Case study with \textcolor{purple}{rare disease} knowledge injection in data synthesis.}
\label{tab:case_study_w}
\end{table*}

\begin{table*}[htbp]
\centering
\resizebox{\linewidth}{!}{
    \begin{tabular}{p{2.5cm}p{20.5cm}}
       \toprule[1pt]
        \multicolumn{2}{c}{\bf Case Study w/o rare disease} \\
       \midrule
       \bf Seed \newline Question 1 & A 35wks POG, USG parameters more or less corresponding to POG. Doppler showing absent end diastolic flow Management:\newline A. Monitor till it becomes reverse diastolic flow or 37 weeks B. Monitor till 37 weeks C. Give steroids and wait for 48 hours D. Plan for immediate termination\\
       \midrule
       \bf Seed \newline Question 2 & A drug 'X' belongs to the anticholinergic drug group. It is primarily used in pre anesthetic medication and also during surgery. Which of the following can be 'X' ?\newline A. Glycopyrrolate B. Pipenzolate methyl bromide C. Isopropamide D. Dicyclomine\\
       \midrule
       \bf Synthetic \newline Question &
       A 32-year-old woman at 33 weeks gestation presents with reduced fetal movements. Ultrasound reveals an estimated fetal weight below the 10th percentile for gestational age. Umbilical artery Doppler shows an elevated systolic/diastolic ratio but preserved end-diastolic flow. Non-stress test is non-reactive. What is the most appropriate next step in management? \newline A. Repeat non-stress test in 24 hours and continue monitoring B. Administer corticosteroids and plan delivery after 48 hours C. Immediate delivery D. Start magnesium sulfate and continue pregnancy until 37 weeks\\
       \midrule 
       \bf Majority \newline Answer \newline Num: 6/8 & 
       \# Step-by-step reasoning:\newline1. **Clinical Presentation**:  - The patient is a 32-year-old woman at 33 weeks gestation.    - She presents with reduced fetal movements, which is concerning for fetal compromise.    - Ultrasound reveals an estimated fetal weight below the 10th percentile for gestational age, indicating fetal growth restriction (FGR).    - Umbilical artery Doppler shows an elevated systolic/diastolic ratio but preserved end-diastolic flow, suggesting chronic placental insufficiency but not acute compromise.    - The non-stress test is non-reactive, which is abnormal and indicates reduced fetal reactivity.  \newline2. **Interpretation of Findings**:    - **Reduced fetal movements** and **estimated fetal weight below the 10th percentile** are strong indicators of fetal growth restriction (FGR).    - **Elevated systolic/diastolic ratio** on umbilical artery Doppler suggests chronic placental insufficiency.    - **Preserved end-diastolic flow** indicates that there is still some placental function, but it is compromised.    - A **non-reactive non-stress test** is abnormal and suggests reduced fetal reactivity, which is concerning for fetal distress.  \newline3. **Management of FGR with Abnormal Non-Stress Test**:    - In cases of FGR with abnormal non-stress tests (e.g., non-reactive), the next step is to assess for acute fetal compromise.    - If there is evidence of acute fetal distress (e.g., abnormal biophysical profile, abnormal Doppler findings), immediate delivery may be indicated.    - In the absence of acute distress, the decision to deliver depends on the gestational age, fetal condition, and maternal factors.  \newline4. **Gestational Age and Decision Making**:    - At 33 weeks gestation, the fetus is considered preterm.    - If there is no evidence of acute fetal distress, delivery is generally not indicated unless there is clear evidence of worsening fetal condition.    - Corticosteroids may be administered to promote fetal lung maturity if delivery is likely within 7 days.  \newline5. **Most Appropriate Next Step**:    - Given the reduced fetal movements, FGR, and abnormal non-stress test, the next step should be to assess for acute fetal distress.    - If there is no evidence of acute distress, **corticosteroids** should be administered to promote fetal lung maturity, and the patient should be monitored closely for further fetal deterioration.    - Delivery is typically not indicated at 33 weeks unless there is clear evidence of acute fetal distress or worsening fetal condition.  \newline6. **Final Decision**:    - **Administer corticosteroids** to promote fetal lung maturity.    - **Plan delivery after 48 hours** if there is no improvement in fetal condition or if there is evidence of acute fetal distress.  \newline\# Final Answer: The final answer is B.\\
       \midrule
       \bf Minority \newline Answer \newline Num: 2/8 & 
       To determine the most appropriate next step in the management of a 32-year-old woman at 33 weeks gestation with reduced fetal movements, let's break this down step-by-step.\newline\#\#\# Step-by-Step Reasoning:\newline1. **Clinical Presentation:**  - The patient is at 33 weeks gestation.  - She presents with reduced fetal movements.  - Ultrasound reveals an estimated fetal weight below the 10th percentile for gestational age (indicating intrauterine growth restriction, IUGR).  - Umbilical artery Doppler shows an elevated systolic/diastolic ratio but preserved end-diastolic flow (suggesting chronic placental insufficiency).  - Non-stress test is non-reactive (indicating poor fetal well-being).\newline2. **Interpretation of Findings:**  - Reduced fetal movements + IUGR + abnormal Doppler + non-reactive NST suggest fetal compromise.  - Elevated systolic/diastolic ratio indicates chronic placental insufficiency, which is often associated with poor outcomes if not addressed promptly.\newline3. **Management Considerations:**  - Intrauterine growth restriction (IUGR) with abnormal Doppler findings and non-reactive NST is concerning for fetal compromise.  - Immediate intervention may be necessary to prevent adverse outcomes such as stillbirth or neonatal morbidity.\newline4. **Options Analysis:**  - **A: Repeat non-stress test in 24 hours and continue monitoring** - This is not appropriate given the presence of non-reactive NST and other concerning findings.  - **B: Administer corticosteroids and plan delivery after 48 hours** - Corticosteroids can be given to promote fetal lung maturity, but immediate delivery might be warranted based on the findings.  - **C: Immediate delivery** - Given the combination of reduced fetal movements, IUGR, abnormal Doppler, and non-reactive NST, immediate delivery is often recommended to prevent further fetal compromise.  - **D: Start magnesium sulfate and continue pregnancy until 37 weeks** - Magnesium sulfate is typically used for neuroprotection in preterm labor, not for managing IUGR with compromised fetal status.\newline5. **Final Decision:**  - The presence of non-reactive NST and IUGR with abnormal Doppler findings suggests significant fetal compromise.  - Immediate delivery is generally recommended in such cases to prevent adverse outcomes.\newline\#\#\# Final Answer: Given the clinical presentation and findings, the most appropriate next step in management is **immediate delivery**. Therefore, the final answer is C.\\
       \bottomrule[1pt]
    \end{tabular}
}
\caption{Case study without \textcolor{purple}{rare disease} knowledge injection in data synthesis.}
\label{tab:case_study_wo}
\end{table*}

To showcase the generated question and the label voting process, we present case studies with and without rare disease knowledge injection in Table~\ref{tab:case_study_w} and Table~\ref{tab:case_study_wo}, respectively.
We present an example of the majority answer and that of a minority answer.

The comparison between Table~\ref{tab:case_study_w} and Table~\ref{tab:case_study_wo} shows the impact of disease‑specific knowledge injection.
When using the rare disease Worster‑Drought Syndrome to retrieve, we obtain the relevant clinical descriptions.
This informed context steers the generation toward a targeted, clinically‑grounded diagnostic question specific to Worster‑Drought Syndrome.
The case demonstrates how our knowledge‑enhanced synthesis pipeline effectively tailors question content to underrepresented medical domains, enhancing both the relevance and factual grounding of the generated data.

\subsection{Failure Mode Analysis}\label{appendix:failure}

\begin{table*}[htbp]
\centering
\resizebox{\linewidth}{!}{
    \begin{tabular}{p{2.5cm}p{20.5cm}}
       \toprule[1pt]
        \multicolumn{2}{c}{\bf Failure Case} \\
       \midrule
       \bf Question & A 55-year-old woman presents with progressive weakness in her right hand and difficulty manipulating small objects over the past 3 months. She also reports mild pain radiating from her neck to her right shoulder. On examination, there is atrophy of the right thenar muscles and decreased sensation over the lateral aspect of the right hand. What is the most likely diagnosis?\\
       \midrule
       \bf Options & A. Carpal tunnel syndrome, B. Cervical radiculopathy (C6-C7), C. Ulnar nerve entrapment at the elbow, D. Thoracic outlet syndrome.\\
       \midrule
       \bf Ground Truth & A. Carpal tunnel syndrome\\
       \midrule
       \bf Voting Results & "A": 0, "B": 8, "C": 0, "D": 0\\
       \midrule
       \bf Response \newline Example & To determine the most likely diagnosis for this patient, let's break it down step-by-step:\newline\#\#\# Step-by-Step Reasoning:\newline1. **Clinical Presentation**: - Age: 55 years old  - Symptoms: Progressive weakness in the right hand, difficulty manipulating small objects, mild pain radiating from the neck to the right shoulder - Physical Exam: Atrophy of the right thenar muscles, decreased sensation over the lateral aspect of the right hand\newline 2. **Key Findings**:  - **Atrophy of the thenar muscles**: This is characteristic of median nerve involvement. - **Decreased sensation over the lateral aspect of the right hand**: The lateral aspect of the hand is innervated by the radial nerve, but the thenar muscles are primarily innervated by the median nerve.  - **Radiating pain from the neck to the shoulder**: This suggests a cervical spine issue rather than a peripheral nerve problem.\newline 3. **Differential Diagnosis**: - **A: \textcolor{red}{Carpal Tunnel Syndrome**: Typically presents with numbness and tingling in the thumb, index, middle, and half of the ring finger, but usually not with thenar muscle atrophy or neck pain}. - **B: Cervical Radiculopathy (C6-C7)**: Commonly presents with weakness in the wrist extensors (extensor carpi radialis brevis, extensor carpi radialis longus, and extensor carpi ulnaris), which can cause difficulty with small object manipulation. It also often involves pain radiating from the neck to the shoulder, and sensory changes in the thumb, index, and middle fingers (C6) or the middle and ring fingers (C7). Thenar muscle atrophy is more commonly associated with C7 radiculopathy. - **C: Ulnar Nerve Entrapment at the Elbow**: Presents with weakness in the hypothenar muscles, adductor pollicis, and interossei, along with decreased sensation in the little finger and half of the ring finger. There is no significant atrophy of the thenar muscles. - **D: Thoracic Outlet Syndrome**: Can present with symptoms similar to those of cervical radiculopathy, including pain, numbness, and weakness in the upper extremity, but typically involves both arms and may be associated with vascular signs like cyanosis or pallor of the hand.\newline4. **Specificity of Symptoms**: - **Thenar muscle atrophy** is strongly suggestive of median nerve involvement, which is often due to compression at the wrist (carpal tunnel syndrome) or cervical spine issues (e.g., C6 or C7 radiculopathy). - **Radiating pain from the neck to the shoulder** points towards a cervical spine origin.  - **Lateral aspect of the hand** is not typically affected by median nerve issues but could be related to radial nerve issues, which are less likely given the presence of thenar muscle atrophy.\newline5. **Conclusion**: - The combination of thenar muscle atrophy, decreased sensation over the lateral aspect of the hand, and radiating pain from the neck to the shoulder is most consistent with **C6-C7 cervical radiculopathy**. - The lateral aspect of the hand is innervated by the radial nerve, which can be affected by C6 radiculopathy, but the presence of thenar muscle atrophy points more strongly to C7 involvement.\newline\#\#\# Final Answer: Based on the clinical presentation and physical exam findings, the most likely diagnosis is **B: Cervical radiculopathy (C6-C7)**.\newline The final answer is B"\\
       \bottomrule[1pt]
    \end{tabular}
}
\caption{Failure case analysis of majority-voted labels. The incorrect parts are highlighted in \textcolor{red}{red}.}
\label{tab:failure_case}
\end{table*}

To examine whether systematic errors in the base model could be amplified by majority voting in our self-labeling pipeline, we analyzed an incorrect case flagged in the physician evaluation to characterize the nature of such failures.

As shown in Table~\ref{tab:failure_case}, the base model’s majority vote (8/8 responses) selected answer B (Cervical radiculopathy).
Examining the reasoning traces revealed that the model incorrectly output that \textit{...Carpal tunnel syndrome typically presents with numbness and tingling in the thumb, index, and middle fingers, but usually not with thenar muscle atrophy or neck pain...}.
However, thenar atrophy is exactly a hallmark of advanced/severe carpal tunnel syndrome.
This erroneous internal knowledge led the model to confidently exclude the correct answer and settle on a plausible but incorrect alternative.

Crucially, the model’s reasoning in this case was internally consistent.
It did not exhibit random guessing or logical contradictions, but rather a coherent chain based on wrong knowledge.
This type of failure represents a systematic knowledge error.

While systematic errors can propagate through majority voting, our analysis shows that they do not undermine the overall training effectiveness for two reasons.
First, our physician evaluation confirms that such failures are extremely rare.
Second, even when incorrect pseudo-labels are used in the self-supervised stage, the subsequent supervised training stage introduces accurate external supervision from real data, which can help correct such errors and refine the model's knowledge boundaries.

\section{Prompt List}\label{appendix:prompt}

The prompt used for data synthesis \textbf{without} medical knowledge, data synthesis \textbf{with} medical knowledge, reasoning data selection, and evaluation are present in Prompt~\ref{prompt:syn_noknowledge},~\ref{prompt:syn_knowledge},~\ref{prompt:select},~\ref{prompt:test}, respectively.

\begin{figure*}[t]
\begin{prompt}[title={Prompt \thetcbcounter: Prompt for Data Synthesis without Medical Knowledge}, label=prompt:syn_noknowledge]
The following reasoning-heavy seed questions are provided:

    Seed Question 1: \{seed\_question1\}

    Seed Question 2: \{seed\_question2\}
\\

Your task is to generate a medical reasoning question following these requirements:

(a) Using the seed question as inspiration, generate a novel and clinically accurate question of comparable difficulty.

(b) Ensure all medical content is factually correct and avoid counterfactual content (e.g., male patients with ovarian cancer)

(c) Questions must be self-contained and require multi-step reasoning, not simple fact recall

(d) Include realistic multiple-choice options that are plausible but only one is correct
\\

\end{prompt}
\end{figure*}

\begin{figure*}[t]
\begin{prompt}[title={Prompt \thetcbcounter: Prompt for Data Synthesis with Medical Knowledge}, label=prompt:syn_knowledge]
The following reasoning-heavy seed questions are provided:
\\
Seed Question 1: \{seed\_question1\}
\\
Seed Question 2: \{seed\_question2\}
\\

MEDICAL KNOWLEDGE CONTEXT:
\\
The following medical knowledge about \{rare\_disease\} can be used:

\{medical\_knowledge\}
\\

Your task is to generate a medical reasoning question following these requirements:
\\
(a) Using the seed question as inspiration, generate a novel and clinically accurate question of comparable difficulty.
\\
(b) Incorporate the MEDICAL KNOWLEDGE about rare disease if it is useful
\\
(c) Ensure all medical content is factually correct and avoid counterfactual content (e.g., male patients with ovarian cancer)
\\
(d) Questions must be self-contained and require multi-step reasoning, not simple fact recall
\\
(e) Include realistic multiple-choice options that are plausible but only one is correct
\\

\end{prompt}
\end{figure*}

\begin{figure*}[t]
\begin{prompt}[title={Prompt \thetcbcounter: Prompt for Reasoning Data Selection}, label=prompt:select]
Determine if this medical question requires:
\\
1. MEMORIZATION (can be answered by recalling a single fact)

2. REASONING (requires multi-step analysis of information)
\\

Key characteristics of reasoning questions:
\\
- Require comparing multiple factors
\\
- Involve interpreting clinical scenarios\
\\
- Need synthesis of information
\\
- Ask "most likely" or "best next step"
\\

Respond EXACTLY with:
\\
<Memorization> - for fact-recall questions
\\
<Reasoning> - for analysis questions
\\

Examples:
\\
Memorization:
\\
"What is the diagnostic criteria for X?"
\\
"What is the first-line treatment for Y?"
\\

Reasoning:
\\
"Given these symptoms A, B and C, what is the most likely diagnosis?"
\\
"How would you manage this patient with X and Y conditions?"
\end{prompt}
\end{figure*}

\begin{figure*}[t]
\begin{prompt}[title={Prompt \thetcbcounter: Prompt for Evaluation}, label=prompt:test]
You are a professional medical expert to answer the \# Question. Please first think step-by-step and then answer the question (only one option can be chosen). Your responses will be used for research purposes only, so please have a definite answer and keep your response complete. Please format the final answer at the end of your response as: The answer is [LETTER]. Example: "The answer is A".
\\
Question:
\{question\}
\end{prompt}
\end{figure*}

\end{document}